# RSI-CB: A Large-Scale Remote Sensing Image Classification Benchmark via Crowdsource Data

Haifeng Li, Xin Dou, Chao Tao, Zhixiang Hou, Jie Chen, Jian Peng, Min Deng, Ling Zhao

School of Geosciences and Info-Physics, Central South University

Abstract: Remote sensing image classification is a fundamental task in remote sensing image processing. In recent years, deep convolutional neural network (DCNN) has seen a breakthrough progress in natural image recognition because of three points: universal approximation ability via DCNN, large-scale database (such as ImageNet), and supercomputing ability powered by GPU. The remote sensing field is still lacking a large-scale benchmark compared to ImageNet and Place2. In this paper, we propose a remote sensing image classification benchmark (RSI-CB) based on massive, scalable, and diverse crowdsource data. Using crowdsource data, such as Open Street Map (OSM) data, ground objects in remote sensing images can be annotated effectively by points of interest, vector data from OSM, or other crowdsource data. The annotated images can be used in remote sensing image classification tasks. Based on this method, we construct a worldwide large-scale benchmark for remote sensing image classification. This benchmark has two sub-datasets with 256 × 256 and 128 × 128 sizes because different DCNNs require different image sizes. The former contains 6 categories with 35 subclasses of more than 24,000 images. The latter contains 6 categories with 45 subclasses of more than 36,000 images. The six categories are agricultural land, construction land and facilities, transportation and facilities, water and water conservancy facilities, woodland, and other lands, and each has several subclasses. This classification system of ground objects is defined according to the national standard of land-use classification in China and is inspired by the hierarchy mechanism of ImageNet. Finally, we conduct many experiments to compare RSI-CB with the SAT-4, SAT-6, and UC-Merced datasets on handcrafted features, such as scale-invariant feature transform, color histogram, local binary patterns, and GIST, and classical DCNN models, such as AlexNet, VGGNet, GoogLeNet, and ResNet. In addition, we show that DCNN models trained by RSI-CB have good performance when transferred to another dataset, that is, UC-Merced, and good generalization ability. Experiments show that RSI-CB is more suitable as a benchmark for the remote sensing image classification task than other benchmarks in the big data era and has potential applications.

# 1 Introduction

Remote sensing image classification is a fundamental task in remote sensing image processing **and there has great meaningful work been done(Cheng et al., 2015; Cheng et al., 2014; Cheng et al., 2016; Han et al., 2015; Yao et al., 2016; Yao et al., 2017)**. Deep convolution neural networks (DCNNs) have been considered a breakthrough technology since AlexNet (Krizhevsky et al., 2012) achieved impressive results in the ImageNet Challenge (Deng et al., 2009) in 2012. DCNN brought computer vision applications to a new era of applications, such as image classification (He et al., 2015a; Krizhevsky et al., 2012; Simonyan and Zisserman, 2014b; Zhou et al., 2016), face recognition (Sun et al., 2013, 2014a, b, 2015; Taigman et al., 2014), video analysis (Donahue et al., 2015; Ji et al., 2013; Simonyan and Zisserman, 2014a; Yan et al., 2014), and object detection (Girshick, 2015; Girshick et al., 2014; Liu et al., 2016a; Redmon et al., 2016; Ren et al., 2015; Zeng et al., 2016). Specifically, ResNet (He et al., 2015a), which was developed by the Microsoft Asian Institute of Visual Computing Group in 2015, achieved a 3.84% error rate in the ImageNet1000 Challenge for the first time, surpassing human performance in the dataset (He et al., 2015b). Inspired by the success of using DCNN on natural image classification, remote sensing experts introduced DCNN to remote sensing image classification and other recognition tasks (Chen et al., 2014; Hu et al., 2015; Nogueira et al., 2017; Penatti et al., 2015; Salberg, 2015), which is the current frontier and highlight of remote sensing image processing.

The key factors of DCNN model's success lie in the following points: universal approximation ability via deep convolutional network, large-scale database (such as ImageNet), and supercomputing ability powered by GPU. DCNN can learn effective feature representation from large-scale training samples, and these features are extremely important for computer vision tasks. Using large training samples is significant in two aspects. (1) Massive samples could help DCNN, with very complex and millions of parameters, avoid over-fitting and obtain a feature expression that is more effective. (2) Large-scale samples could help fill or approximate the entirety of the sample space as much as possible, which is an important factor for DCNN's generalization ability in some real-world applications.

However, the remote sensing image field is lacking a large-scale benchmark. In earlier times, remote sensing image benchmarks, such as, NLCD 1992 (Vogelmann, 2001), NLCD 2001 (Homer et al., 2007), and NLCD 2006 (Fry et al., 2011; Xian et al., 2011), were characterized by low spatial resolution and single target classification. These benchmarks were composed of land cover images with 30-m spatial resolution, which were not suitable for remote sensing image classification because distinguishing different objects in such a low spatial resolution image was extremely difficult. Recently, researchers have begun to build high-resolution remote sensing image benchmarks for special tasks. For instance, a road detection benchmark was constructed with a 1.2-m spatial resolution that covers approximately 600 square kilometers of area (Mnih and Hinton, 2010). A vegetation coverage identifying benchmark was proposed using aerial images with 1-m spatial resolution (Basu et al., 2014). In recent years, remote sensing benchmark has been moving toward high spatial resolution and complexity objective classification. For example, ISPRS provided a dataset with aerial imageries of homes, roads, vegetation, artificial ground, and other objects, as well as those of building reconstruction (Rottensteiner et al., 2012). UC-Merced (Yang and Newsam, 2010) is a famous satellite imagery database for scene classification. It is made up of urban area images from the United States Geological Survey (USGS). UC-Merced includes 21 categories and 2,100 256 × 256 images with 0.3-m spatial resolution. SAT-4 and SAT-6 are two other big benchmarks for remote sensing classification collected by the National Agriculture Imagery Program (NAIP) (Basu et al., 2015). SAT-4 and SAT-6 consist of six categories, namely, bare soil, vegetation, grassland, road, house, and water body in California. SAT-4 and SAT-6 have 500,000 and 405,000 images, respectively. The sizes of the image patches are 28 × 28. Terrapattern (Levin, 2016), which was constructed by Carnegie Mellon University, can search for similar objects according to given objects within the specified area (currently includes New York, San Francisco, Pittsburgh, Detroit, Miami, Austin, and Berlin, Germany). The search results can be marked on the map, which provides a new method for constructing a remote sensing image dataset. The aerial image dataset (AID) (Xia et al., 2016) contains 10,000 aerial image patches with 0.5-m to 8-m spatial resolutions. All 10,000 image patches were mapped into 30 categories, such as airport, desert, farmland, school, river, and other objects. NWPU-RESISC45 (Cheng et al., 2017), which was built by the Northwestern Polytechnic University, has 45 object categories with 31,500 images of 0.2-m to 3-m spatial resolutions.

A large-scale benchmark is critical to remote sensing experts for improving their models and algorithms because deep learning methods have governed imaging related tasks in the big data era. However, building high-quality and larger-scale benchmarks is challenging because (1) a remote sensing image has numerous complex objects rather than simple objects of one nature, (2) objects in remote sensing are in global scale, and (3) remote sensing images are affected by several factors, such as the camera's perspective, weather condition, and solar altitude angle. Crowdsource geographic data are open geospatial data provided by the public or related institutions through the Internet (Heipke, 2010). Crowdsource geographic data sources are large, rich in information, diverse in categories, low cost, and real time (Rice M T, 2012). Hence, using crowdsource geographic data as annotations for remote sensing images is a potential approach to building high-quality and larger-scale benchmarks.

This paper presents a method based on crowdsource geographic data for building a remote sensing image benchmark. We select points of interests (POIs) from the Open StreetMap (OSM) (Haklay and Weber, 2008) of different places. Then, we align the POIs with different temporal remote sensing images downloaded from the Bing Map server according to the geo-coordinate. We call this remote sensing image classification benchmark (RSI-CB[1]), which uses POIs from different countries. According to the distribution density of POIs, we select high-density areas in Beijing, Shanghai, Hong Kong, Guangzhou, Hainan, Fujian, and other provinces in China; New York, Washington, Los Angeles, Chicago, and other cities in the US; Tokyo, Osaka, Kobe, and other cities in Japan; Paris, Nice, and other cities in France; Ottawa, Toronto, and other cities in Canada; and Moscow, St. Petersburg, and other cities in Russia. Other countries and their regions that have strict hierarchical systems are also included. The remote sensing images are downloaded from Google Earth and Bing Maps with 0.22–3-m spatial resolutions. Two subsets with varied image sizes of 256 × 256 and 128 × 128 are employed for fitting different DCNN models. We call the former RSI-CB256 and the latter RSI-CB128. The objects have 45 and 35 categories, respectively, including more than 60,000 images. RSI-CB256 contains 6 categories with 35 subclasses of more than 24,000 images. The RSI-CB128 contains 6 categories with 45 subclasses of more than 36,000 images. Inspired by ImageNet (Deng et al., 2009), we combine the hierarchical grading mechanism of China's land-use classification standard (Chen Baiming, 2007) to satisfy the diversity and comprehensive requirements of the object class further for constructing a generalized benchmark. Currently, RSI-CB has six categories, namely, agricultural land, construction land and facilities, transportation and facilities, water and water conservancy facilities, woodland, and other land uses.

The main contributions of this paper are as follows:

(1). We propose a crowdsource data-based method to build a RSI-CB. The crowdsource data in our method is a high-precision supervisor. Traditional methods require a significant amount of manual work; thus, they are less efficient and time-consuming. Using crowdsource data as a supervisor facilitates machine self-learning through the Internet. Moreover, the size of the benchmark sample could be infinite both in amount and variety. In addition, crowdsource data are basic data sources in the big data era and are updated rapidly. Therefore, the remote sensing benchmark constructed using crowdsource data can possibly continue to expand in terms of diversity, quantity, and robustness of samples. Consequently, our method can potentially realize weak supervised learning further for remote sensing image.

[1]This benchmark can be downloaded from https://github.com/lehaifeng/RSI-CB.

(2). Based on the above method, we construct a global -scale RSI-CB. Considering the different image size requirements of the DCNN, we construct two datasets of 256 × 256 and 128 × 128 pixel sizes (RSI-CB256 and RSI-CB128, respectively) with 0.3–3-m spatial resolutions. The former contains 35 categories and more than 24,000 images. The latter contains 45 categories and more than 36,000 images. We establish a strict object category system according to the national standard of land-use classification in China and the hierarchical grading mechanism of ImageNet. The six categories are agricultural land, construction land and facilities, transportation and facilities, water and water conservancy facilities, woodland, and other land.

(3). We conduct various experiments to compare RSI-CB with SAT-4, SAT-6, and UC-Merced datasets on handcrafted features, such as scale-invariant feature transform (SIFT) (Lowe, 2004), color histogram indexing (CH) (Swain and Ballard, 1991), local binary patterns (LBP) (Ojala et al., 2002), GIST (Oliva and Torralba, 2001), and classical DCNN models, such as AlexNet (Krizhevsky et al., 2012), VGGNet (Simonyan and Zisserman, 2014b), GoogLeNet (Szegedy et al., 2014), and ResNet (He et al., 2015a). In addition, we demonstrate that DCNN models trained by RSI-CB have good performance when transferred to other datasets, that is, UC-Merced, and good generalization ability. The experiments show that RSI-CB is a more suitable benchmark for remote sensing image classification than other benchmarks in the big data era, and has potential applications.

The rest of this paper is organized as follows. Section 2 reviews several related remote sensing image benchmarks. Section 3 describes the basic requirements of a remote sensing image benchmark for deep learning and the crowdsource data-based method for building a remote sensing image benchmark. Section 4 presents the analysis of the properties of RSI-CB on geographical distribution, category hierarchy, and statistical distribution and the comparison of RSI-CB with other remote sensing benchmarks. Section 5 discusses the results of the tests on the classification performance using handcrafted feature methods and classic DCNN models on RSI-CB and other benchmarks. Section 6 concludes the paper and presents future directions.

## 2 Related Works

This section introduces several related remote sensing image datasets and compares the important aspects of an image database, such as object class, number of images, spatial resolution, size of images, and visual diversity.

### 2.1 NLCD

The National Land Cover Database (NLCD) includes three subsets, namely, NlCD1992 (Vogelmann, 2001), NLCD2001 (Homer et al., 2007), and NLCD2006 (Fry et al., 2011; Xian et al., 2011). NLCD1992 was the first US land cover database with a 30-m spatial resolution. NLCD2001 added land cover data in three regions (i.e., Alaska, Hawaii, and Puerto Rico) based on NLCD1992. The concept of the database is introduced into maps by incorporating the percentage of urban impervious surface and the percentage of forest cover and improving the land cover classification method. NLCD2006 is a 30-m spatial resolution database with images from Landsat7 and Landsat5, which inherited the NLCD2001. On this basis, NLCD2006 added the change data of land cover and impervious surface between 2001 and 2006.

NLCD datasets are used mainly for the US land cover classification and change detection, which is different from recent classification datasets. NLCD characterizes objects in pixels. These objects include the open category of water, shrubs, grassland, and so on. Identifying small-scale objects is very difficult because of the benchmark's low spatial resolution. Therefore, the database is applicable only to large-scale land-use identification and useless for smaller but important objects, such as traffic transport, water conservancy facilities, grassland, and construction land.

### 2.2 UC-Merced

UC-Merced was built from the USGS. UC-Merced (Yang and Newsam, 2010) is a remote sensing image dataset of 256 × 256 pixel size with a spatial resolution of 0.3 m per pixel. It has 21 categories, including overpass, plane, baseball, beach, building, dense residential, forest, freeway, golf course, harbor, and other objects. Each type of land has 100 images, for a total of 2,100 images.

UC-Merced has geography objects with different textures, colors, and shapes. It has high spatial resolution and is widely used in remote sensing image classification (Cui, 2016; Liu et al., 2016b; Wu et al., 2016; Zhao et al., 2016a; Zhao et al., 2016b; Zhu et al., 2016). In this database, the number of images in each category is small and is not suitable for the distribution of overall characteristics for actual geography objects.

### 2.3 SAT-4 and SAT-6

SAT-4 (Basu et al., 2015) and SAT-6 (Basu et al., 2015) datasets are collected by the NAIP, and their aerial images are obtained from the rural, urban, jungle, mountain, water, and agriculture areas in California. SAT-4 and SAT-6 datasets contain large data, including 500,000 and 405,000 28 × 28 annotated image blocks. They have six categories, including bare soil, vegetation, grass, roads, houses, and water. SAT-4 and SAT-6 have the following disadvantages:

(1) They only have six categories, which cannot achieve the purpose of remote sensing image recognition in practical applications.

(2) The image block size is very small. Hence, details of the internal image are insufficient. The small block size cannot reflect the complex distribution of features completely. The image size may contain only the local features of objects.

### 2.4 AID

The AID (Xia et al., 2016) remote sensing image dataset was built by the Wuhan University. Its images were taken from Google Earth. It contains 10,000 images with 30 categories, and each category contains approximately 330 image blocks with spatial resolutions ranging from 0.5 m to 8 m per pixel. The image blocks used are 600 × 600 sized pixels because of the different resolution of the features.

AID has three significant features. (1) It has strong diversity, which is evident in its various imaging angles, shapes, sizes, colors, surrounding environment, and so on. (2) The difference between several scene objects, such as schools and intensive residential areas is small; they are mostly buildings with similar background. These factors potentially improve the difficulty of classification. (3) AID is large scale compared with other existing datasets, such as UC-Merced datasets, which are widely used. Although a certain overlap exists between AID and UC-Merced in terms of categories, AID is approximately five times than that of UC-Merced in amount. However, AID has two disadvantages. First, AID does not have an efficient database construction method, and it has no set of forming system for developing database categories with more geographical and practical values. Second, the AID database scale is not very large.

### 2.5 NWPU-RESISC45

NWPU-RESISC45 (Cheng et al., 2017), which was built by the Northwestern Polytechnic University, was taken from Google Earth with a spatial resolution of 0.2 m to 30 m per pixel. It has 45 categories and 31,500 256 × 256 images, which are mainly used for remote sensing image scene classification.

In addition to continuously maintaining object diversity based on AID, the NWPU-RESISC45 dataset is characterized by improved image scale. Common object types are considered in the selection of categories and through “OBIA,” “GEOBIA,” “geographic image retrieval,” or other keyword searches to determine the final 45 categories, which indicates that the NWPU-RESISC45 dataset pay more attention to objects with geographical significance and application value. However, the disadvantage lies in the inefficiency of the NWPU-RESISC45 dataset in constructing the remote sensing image database, particularly for not using existing geographic crowdsource data as auxiliary information.

## 3 Remote Sensing Image Labeling Method Based on Crowdsource Data

### 3.1 Basic Requirements for Remote Sensing Image Benchmark Using Deep Learning

Deep learning models, such as DCNN, have seen breakthroughs in various tasks, such as image tracking and scene understanding. DCNN models are very complex and have millions of parameters; hence, they easily overfit on small benchmarks (Rezić, 2011). From the learning theory perspective, DCNN models have high VC dimension, which results in complex and diverse samples (Rezić, 2011). From the optimization algorithms' perspective, DCNN often employs the stochastic gradient descent (SGD) algorithm (Rezić, 2011). SGD algorithm results in very little change in each neuron's parameter when a lower learning rate is used and the sample complexity is limited. Therefore, we argue that the RSI-CB for DCNN models should be governed by the following factors:

1) Scale of benchmark

DCNN models have a strong learning ability and the ability to approximate any function, which are combined with large data to describe the inherent characteristics of large data in a distribution better. Moreover, the image classification effect is related to the depth and width of the network models, which correspond to network complexity, and a more complex network model indicates more training parameters, which require more image training samples. In addition, when sample data is insufficient, sampling error cannot be ignored even if the network is simple. When sample features do not fit well with the distribution of actual features, the knowledge learnt by the model is insufficient as well, leading to unsatisfactory robustness and generalization.

2) Object diversity

A strong distinction exists within classes. In ensuring large-scale data, data should be representative, causing the learning model to learn not only the unique characteristics. Therefore, the same type of objects (such as in aircrafts) should have different sizes, shapes, perspectives, backgrounds, light intensities, colors, and other conditions that diversify objects within a class. Our images come from Google Earth and Bing Maps, so images come from different sensors. To improve class diversity further, the selection within categories should be based on the diversity requirements mentioned above, which can train a network model with high robustness and stronger generalization.

3) Category differences

In addition to the idea that massive representative training data can learn more visual features, the difference between classes is also one of the determinants of image classification accuracy. In dataset construction, if the difference between classes is large, the probability of independent distribution of feature interval for each category is higher and the similarity between classes leads to a higher overlapping ratio of various features. Therefore, a feasible method is to increase the number of images for these categories, which can cause the inter-class feature response interval to have a higher probability of independence distribution to improve the accuracy of image classification further.

According to these requirements, the construction principles for RSI-CB are as follows:

a) Each category is rich with data. The average size of 128 × 128 includes approximately 800 image patches per category, and the average size of 256 ×256 includes approximately 690 patches per category.

b) The object categories are the combination of the category system of the national land-use classification standard in China and the hierarchy mechanism of ImageNet. The level of each category aims to increase the diversity and comprehensiveness of the benchmark further.

c) Land use has various types, which are built according to the Chinese land-use classification standard with a significant difference between classes. These classes have 45 128 × 128 categories and 35 256 × 256 categories.

d) The identification of the entire system is high, which can avoid ambiguous objects and improve image quality.

e) Each class has different imaging angles, sizes, shapes, colors to increase sample diversity, which could improve the model generalization performance and robustness.

f) The RSI-CB selects the images that come from major cities worldwide and the surrounding areas considering the balance of spatial distribution for the selected images.

### 3.2 Remote Sensing Image Labeling Method Based on Crowdsource Geographic Data

Crowdsource geographic data representation is achieved through the GPS route data (such as the OSM used in this paper), map data edited by user collaboration plan (such as Wikimapia), various social networking site data (such as Twitter and Facebook), and the POI data that users check in. These data need to be processed to form standardized geographic information. The crowdsource geographic data labeled from non-professional people are highly real time, has high spread speed, rich in information, low cost, and large in quantity compared with those from traditional geographic information collection and updating methods. It has become the research focus of international geography information science. However, the unbalance of quality and density distribution, redundant and incomplete of data, and the lack of uniform norms and other defects for such kind of data are the main drawbacks for crowdsource use (Haklay, 2010).

The construction of the RSI-CB mainly uses two types of data, namely, POI in OSM and remote sensing image data. OSM elements include layers, nodes, and ways. In the POI data, we define the positions of points for its space location, which are stored as latitude and longitude and the attribute information for points. POIs are selected from different countries and regions worldwide. Remote sensing images are obtained from Google Earth and Bing Maps. The spatial resolution we used ranges from 0.22 to 3 m per pixel. As shown in Table 1, the left side of the table is the category of POI objects, and the right side is the corresponding subclass under the category. The core ideas to construct RSI-CB can be summarized as follows:

(1) Registration overlay of high-resolution remote sensing images and POI data, making sure that the actual targets in images correspond correctly with the POI categories.

(2) POI data screening. The screening includes the deletion of wrong annotations, removal of non-conformance objects which indicates no intersection with national land-use classification standards in China (see Section 4.2), and deletion of objects which are not obvious (see section below for detailed discussion).

(3) Cropping fixed-size remote sensing image blocks according to the POI data, traversing all POI, taking it as the center to crop remote sensing images with fixed sizes of 128 × 128 and 256 × 256, and then integrating the same category of image blocks to build the final benchmark.

Figure 1 shows the process flow of constructing RSI-CB. Figure 2 and Figure 3 show the superposition effect of the POI data and image.

Table 1. POI data

| | |
|---|---|
| **amenity** | arts_centre、atm、bank、bar、bench、bicycle_parking、bicycle_rental、fountain... |
| **barrier** | bollard、gate、block... |
| **building** | apartments、building... |
| **emergency** | fire_hydrant... |
| **highway** | bus_stop、crossing、motorway_junction... |
| **historic** | memorial、monument... |
| **landuse** | park、sports_centre、swimming_pool... |
| **leisure** | park、picnic_table、playground、swimming_pool... |
| **man_made** | antenna、flagpole、monitoring_station、tower... |
| **natural** | peak、tree... |
| **office** | accountant、company... |

| **public_tran sport** | platform、station... |
|---|---|
| **railway** | station、subway_entrance、ventilation_shaft... |
| **shop** | alcohol、antiques、art、books、clothes、convenience、hairdresser... |
| **sport** | gym、yoga... |
| **tourism** | artwork、gallery、hotel、museum... |

Raw data of POI and high resolution remote sensing image

↓

POI data filtering

↓

Find the intersection of land classification standard and POI data, filtering the POI data which outside the intersection.

↓

Based on the geographic coordinate information of each POI after filtering, taking each of the POI as the center and cut the image blocks of the preset size from the remote sensing image.

↓

Based on the image block and the feature type of each POI after filtering, the image blocks and the object types are determined.

Figure 1: Process flow of constructing RSI-CB

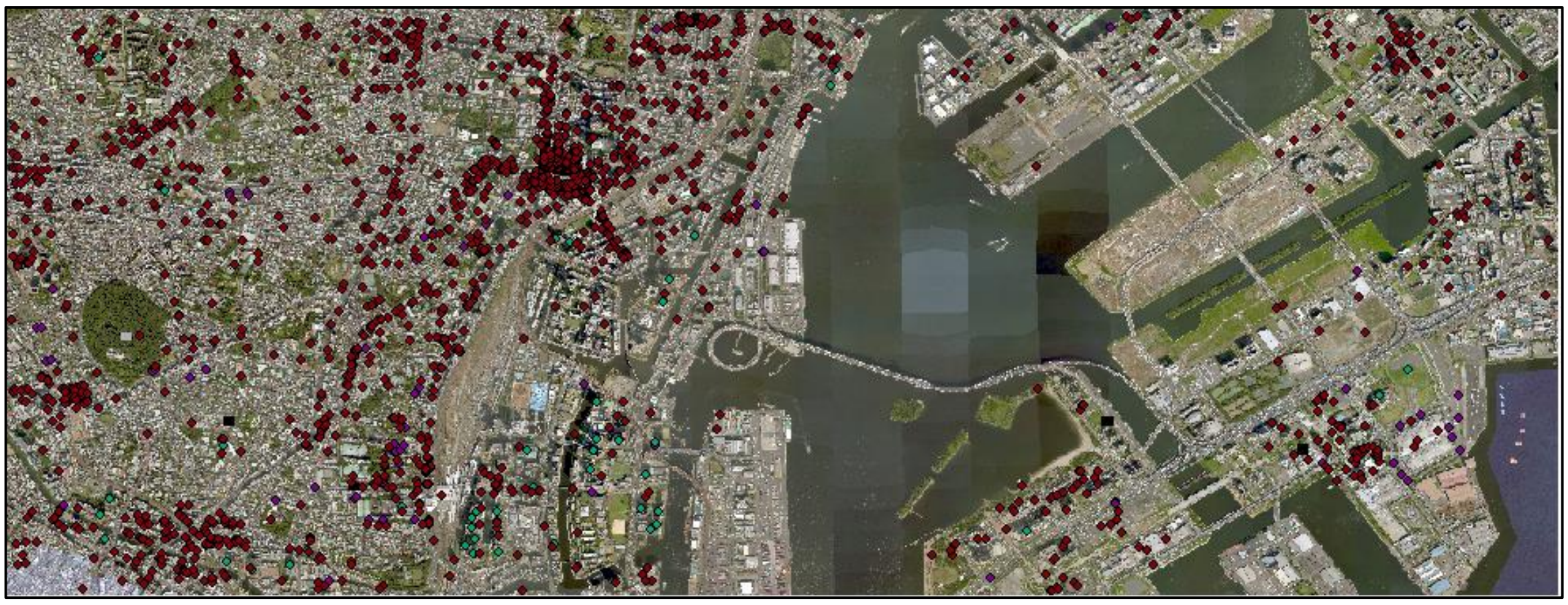

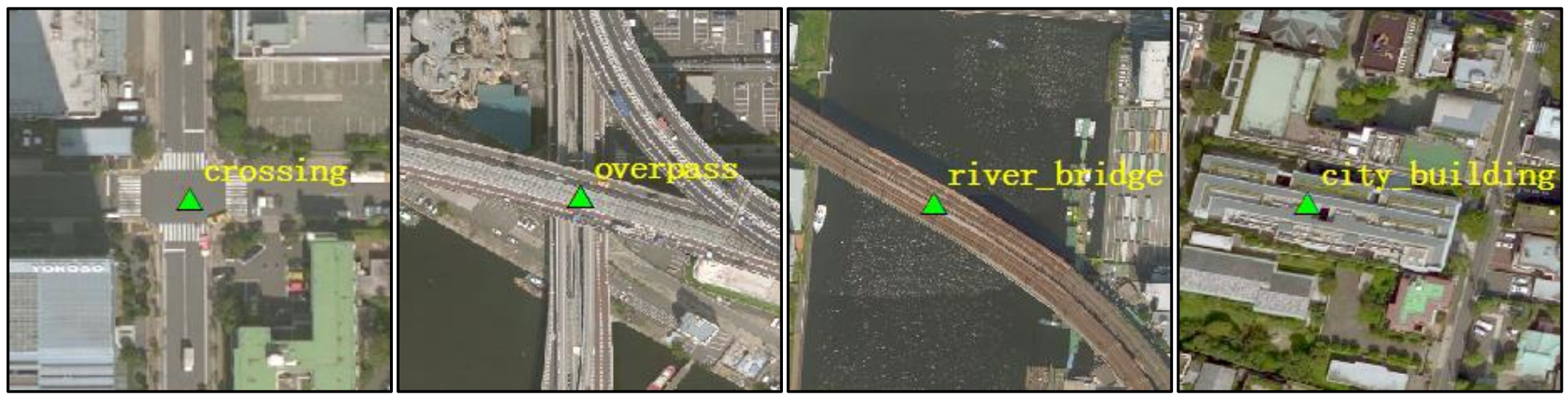

a) crossing b) overpass c) river_bridge d) city_building

Figure 2: Superposition effect of the POI data and image. The four small maps are the categories of crossing, overpass, river_bridge, and city_building, respectively, which are used as the center points for cropping remote sensing images.

However, the uneven coverage of POI data results in different density distributions as well as frequent category updates and incorrect phenomenon labeling by artificial methods. Therefore, using POI to construct a dataset leads to two obvious problems. In this stage, we solve these two problems manually.

1) Several points stacked in a small area, with these points as the center to obtain image blocks (256 × 256 and 128 × 128) with higher overlapping ratios. For this phenomenon, data screening follows the principle that the area of such category is dominant and the recognition degree is high.

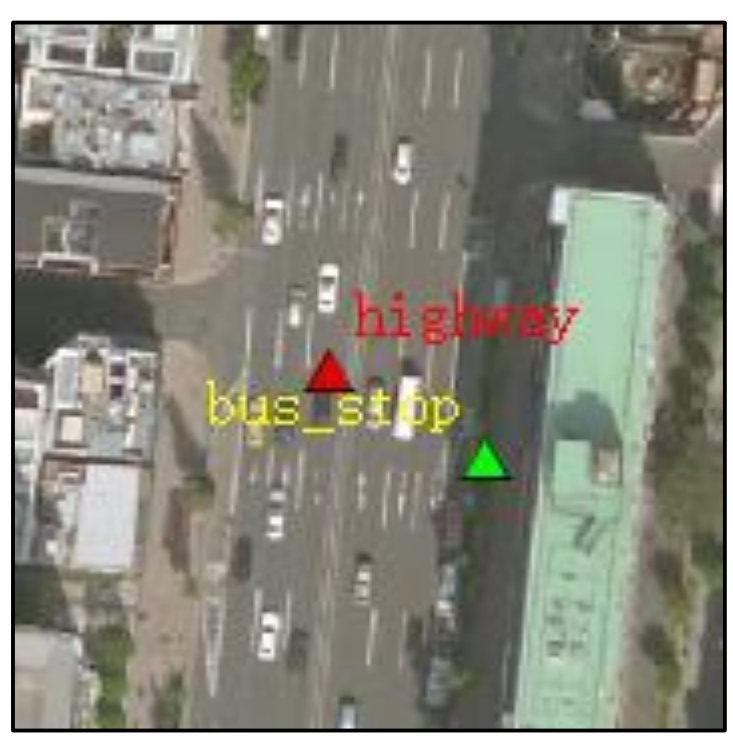

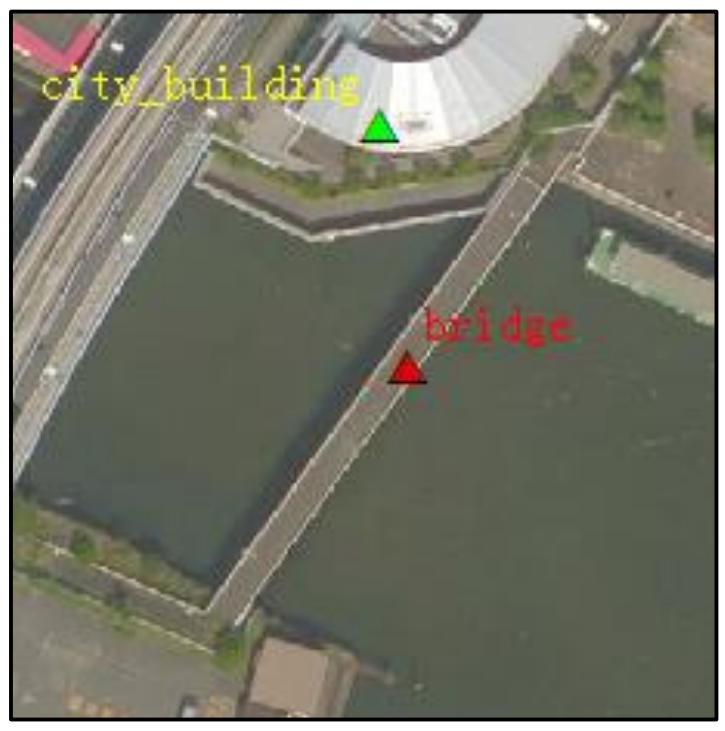

Figure 3 : Multi-labeling in a small candidate area. Red annotations are the retention categories, and the yellow annotations should be deleted. The left side of the candidate area shows “highway” and “bus_stop,” and “bus_stop” should be removed because of its small size and unclear recognition. The right side of the candidate area shows “bridge” and “city_building” should be removed because it’s not salience and it’s smaller size than the bridge, however maybe not for another picture if this city_building is dominant.

2) For some POI data, the names of the objects are not exactly matched with that of the remote sensing image objects. For this phenomenon, delete the POI data directly.

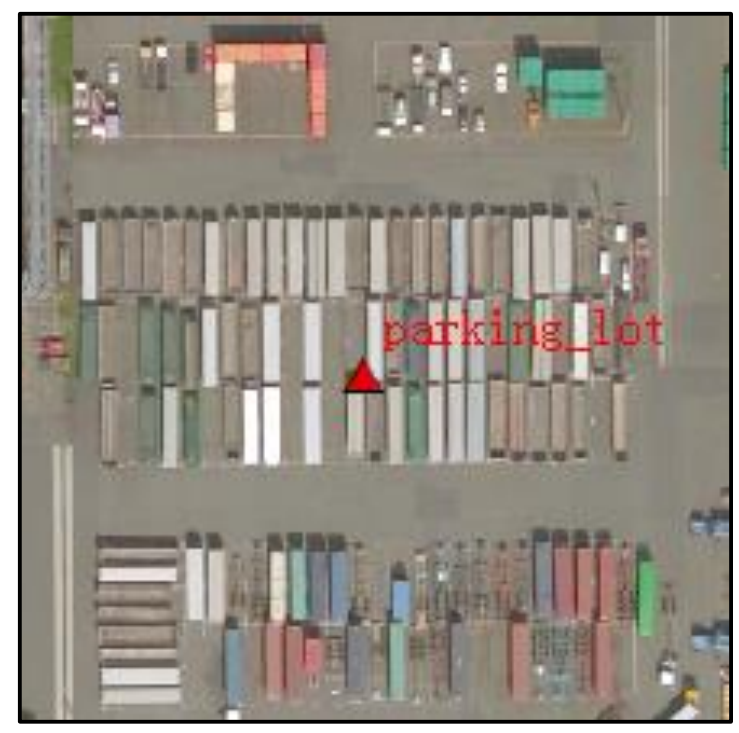

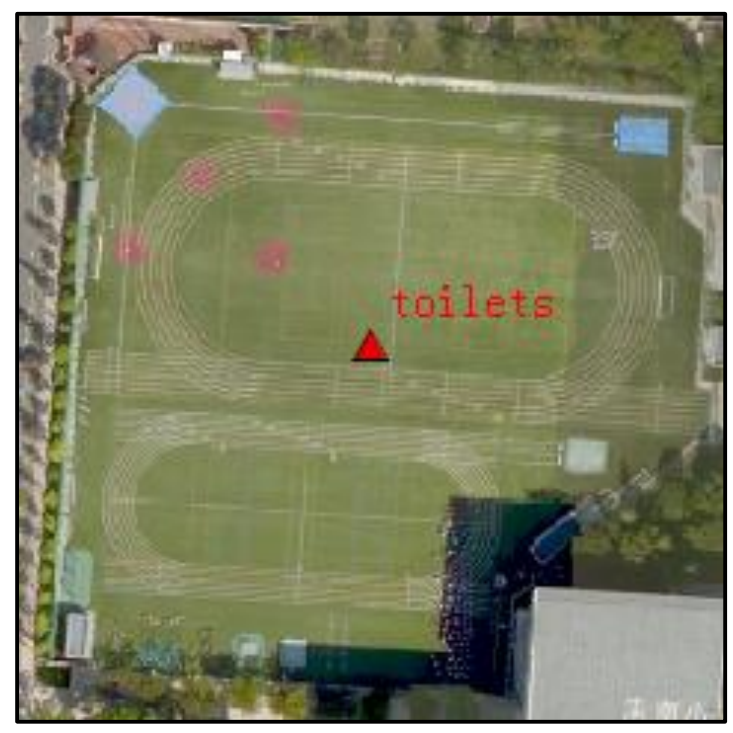

Figure 4: Candidate area error labeling phenomenon. The red label is the wrong category for images. The left area should be labeled as "container" not "parking_lot." The right area should be labeled "artificial_grassland" not "toilets."

# 4 RSI-CB Statistical Analysis

## 4.1 Geographical Distribution of RSI-CB

OSM is a collaboration program of crowdsource map to create a free map. Everyone worldwide could contribute to OSM. Mass volunteers from 230 institutions and countries uploaded data to the OSM every day. Hence, the data from OSM are large scale and widely geographically distributed. Those OSM data can be used as annotation sources for sustainable labeling of global remote sensing images. For instance, Figure 5 shows that the average number of users who label OSM data daily are from the US, Australia, Russia, Europe, and China and had the highest number of labels, and the number of users in Africa were relatively small. Figure 6 shows the distribution of POI data worldwide. Russia had POIs of up to 21% due to its extensive land area and local taxi companies and travel agencies contributing their data. Germany followed with 18.3%. The UK, France, US, and other countries are also taking a greater proportion. Hence, we can yield balanced geo-labels for remote sensing images by controlling the ratio of POIs from different regions worldwide.

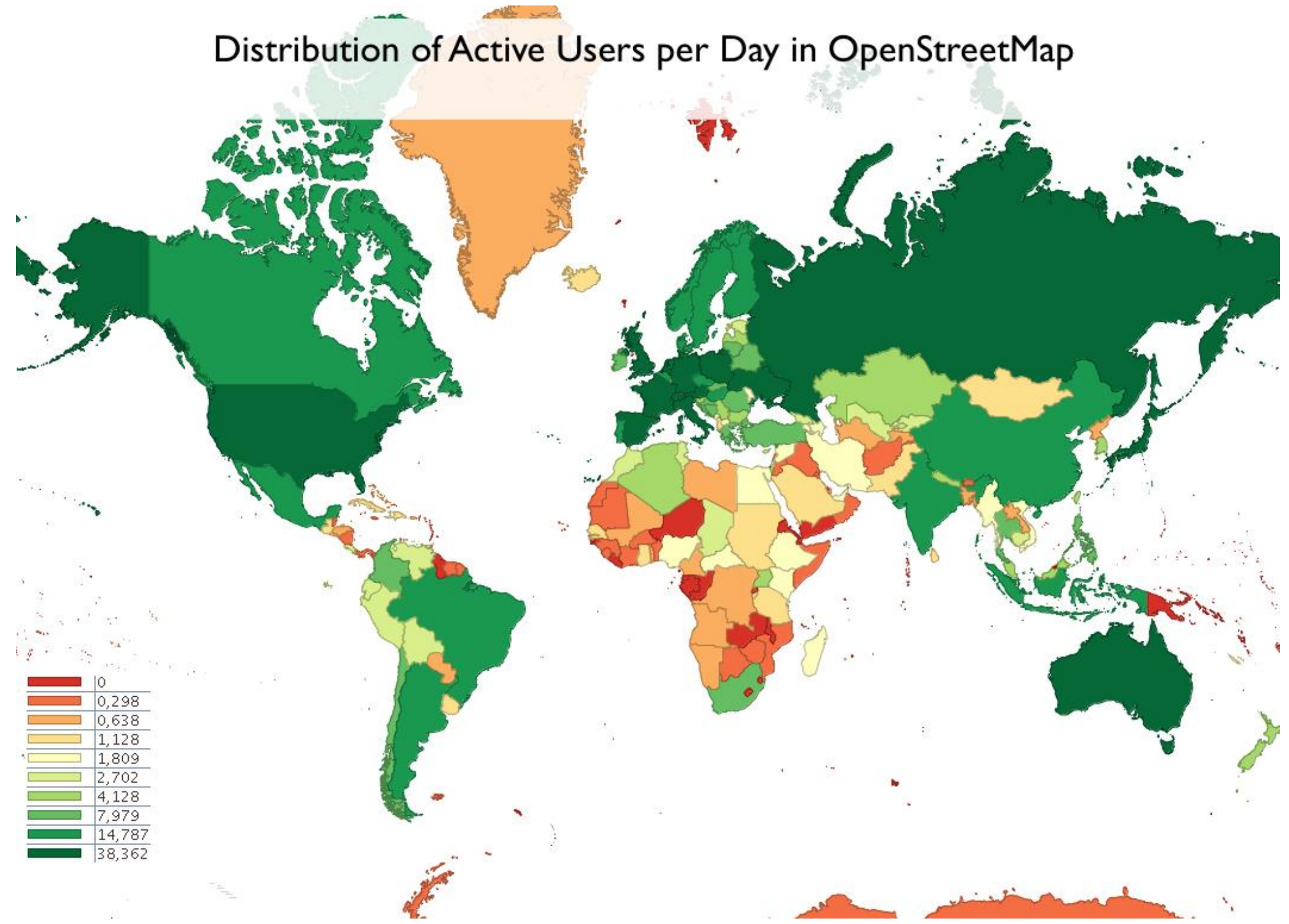

Figure 5: Distribution of active users contributing to OSM data daily.

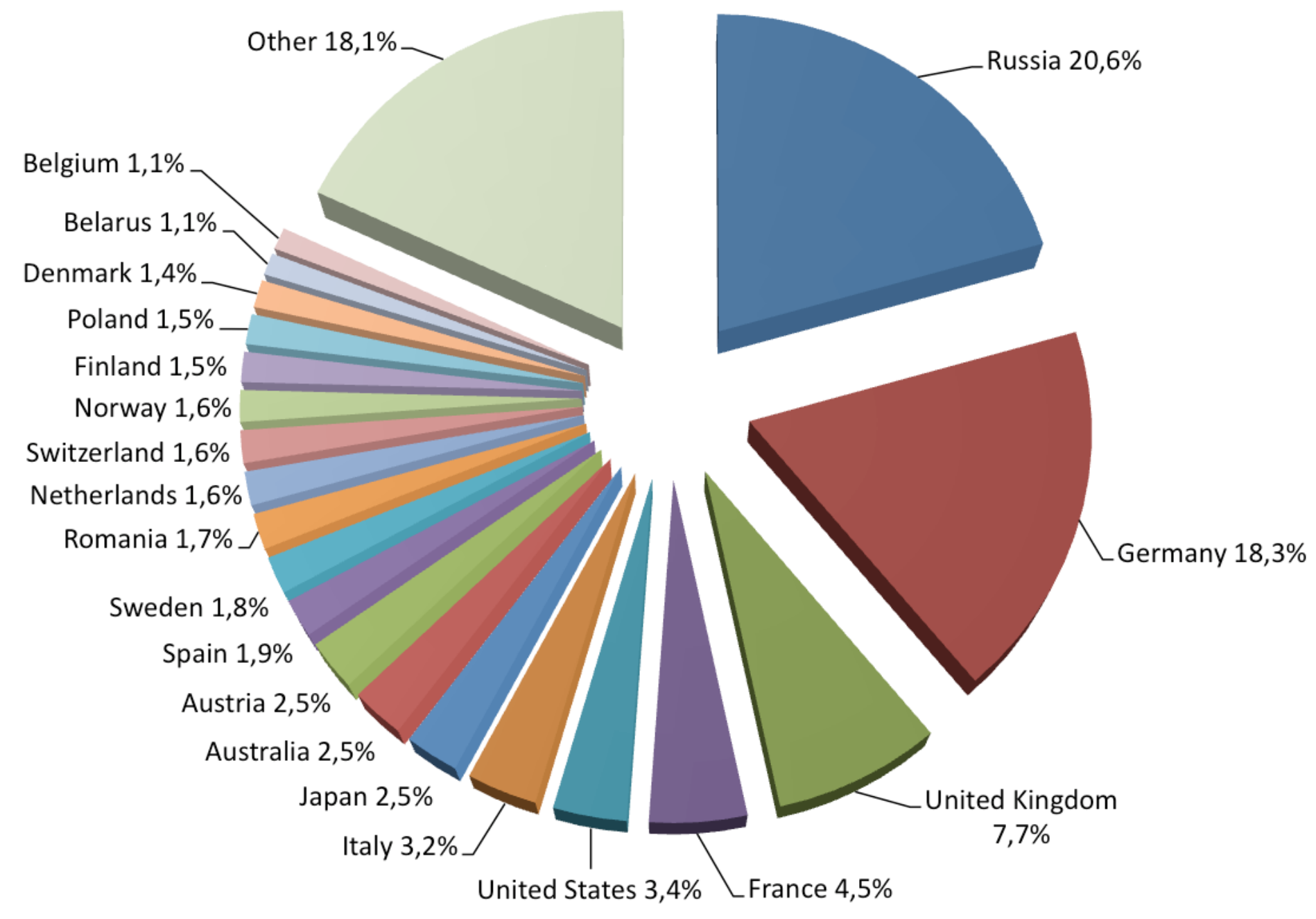

Figure 6: Distribution of POI data worldwide.

Section 3.1 presents the requirements for constructing database. Hence, to train a more robust and generalized classification model, besides the requirements that the database itself should be large, another key points also make sense. **There are four factors to be considered First, the database is designed to be used by researchers around the world, and the selected areas should contain as many cities as possible and with the characteristics of the regions among different categories; second, to meet the requirements of sample diversity, images** **of RSI-CB come from various regions of the world which categories have different shapes, sizes, color characteristics, but also have different intensity of light, shadow conditions and background; third, according to the distribution of the crowdsource data, the selected cities have a large number of data distribution; fourth, selecting the regions which its remote sensing image should have high spatial resolution Taking into account the above four factors,** we select the POI data distributed in major countries and their cities and regions around the world, such as Beijing, Shanghai, Hong Kong, Guangzhou, Hainan, Fujian, and other provinces in China; New York, Washington, Los Angeles, Chicago, and other cities in the US; London, Liverpool, Manchester, and other cities in the UK; Berlin, Cologne, Hamburg, Munich, and other cities in Germany; Osaka, Kobe, and other cities in Japan; Paris, Nice, and other cities in France; Ottawa, Toronto, and other cities in Canada; Rio de Janeiro, Sao Paulo, and other cities in Brazil; as well as other countries and their regions worldwide.

Figure 7 shows the geographical and quantitative distribution of the selected POI data in major countries and regions. The darker area indicates the greater amount of POI data selected, and the blue area indicates that RSI-CB has not yet been selected. Among them, the US, China, and the UK occupy the largest amount of data, including categories of agricultural land, construction land and facilities, transportation and facilities, as well as other large categories. At the same time, the Brazilian woodland, Egyptian desert, and Canadian snow have a different number of distributions.

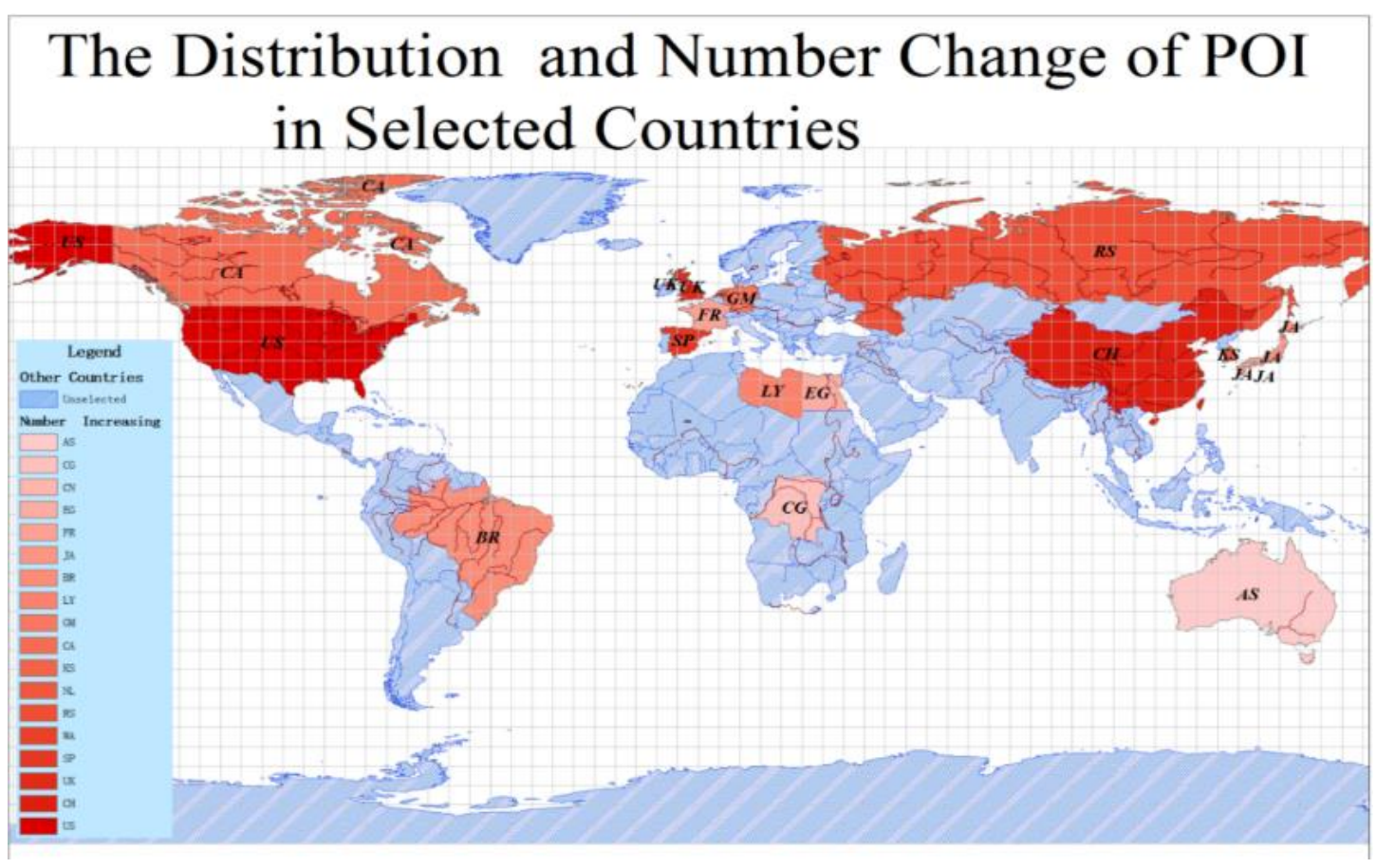

Figure 7: Change in the distribution and number POI in selected countries.

## 4.2 Hierarchy for RSI-CB Categories

The purpose of remote sensing image classification is to extract important and significant feature types in images (Melgani and Bruzzone, 2004). However, given the complexity of the background environment of remote sensing images, diversity of the feature categories, and other external conditions of the objects such as shape, size, color, and other factors, determining the categories of objects, classifying and ensuring the rationality and practical significance of the categories are key points in constructing remote sensing image benchmark. Hence, to meet the diversity within and among classes, we have conducted two methods to establish our classification criteria:

(1). We analyzed the existing categories of OSM and the classification criteria of land use in China and select the common categories among them as the preferred classes.

(2). We deleted the data that do not meet the basic requirements of the benchmark as well as the principles of constructing RSI-CB in Section 3.1.

Table 2 shows the correspondence between China's land-use classification standard and OSM categories. The vertical side of the table is the major class for OSM, and the horizontal side is the land classification standard. Some classes of OSM correspond to several categories of land classification standard and vice versa.

Table 2 The intersecting categories between China land classification standard and POI

| **Land Classification**<br>**OSM Classification** | **Residential** | **Shopping, business** | **Industrial, manufacturing** | **Social, institutional, infrastructure-related** | **Travel or movement** | **Leisure** | **Natural resources-related** | **No human activity or unclassifiable** | **Mass assembly of people** |
|---|---|---|---|---|---|---|---|---|---|
| **amenity** | | √ | | √ | | | | | |
| **barrier** | | | | | | | | | |
| **building** | √ | | | √ | | | | | |
| **emergency** | √ | | | | | | | | |
| **highway** | | | | | √ | | | | |
| **historic** | | | | | | | | | √ |
| **landuse** | √ | | | | | | | | |
| **leisure** | | | | | | √ | | | |
| **man_made** | | | √ | | | | | | |
| **natural** | | | | | | | √ | | |
| **office** | | | | | | √ | | | |
| **public_transport** | | | | | √ | | | | |
| **railway** | | | | | √ | | | | |
| **shop** | | √ | | | | | | | |
| **sport** | | | | | | √ | | | |
| **tourism** | | | | | | √ | √ | | |

**Note: √ indicating contain both of them**

RSI-CB uses Google Earth and Bing Maps as image sources with 0.22–3m spatial resolution. We built RSI-CB128 (128 × 128 pixels) and RSI-CB256 (256 × 256 pixels) to make sure that different categories can adapt to the selected image size and to meet the requirements of the input image for different depth network models. RSI-CB128 and RSI-CB256 have many repetitions in the categories. The differences were mainly in the complexity of the background and the area size of objects occupying the images. Figure 8 shows some subclasses in the categories.

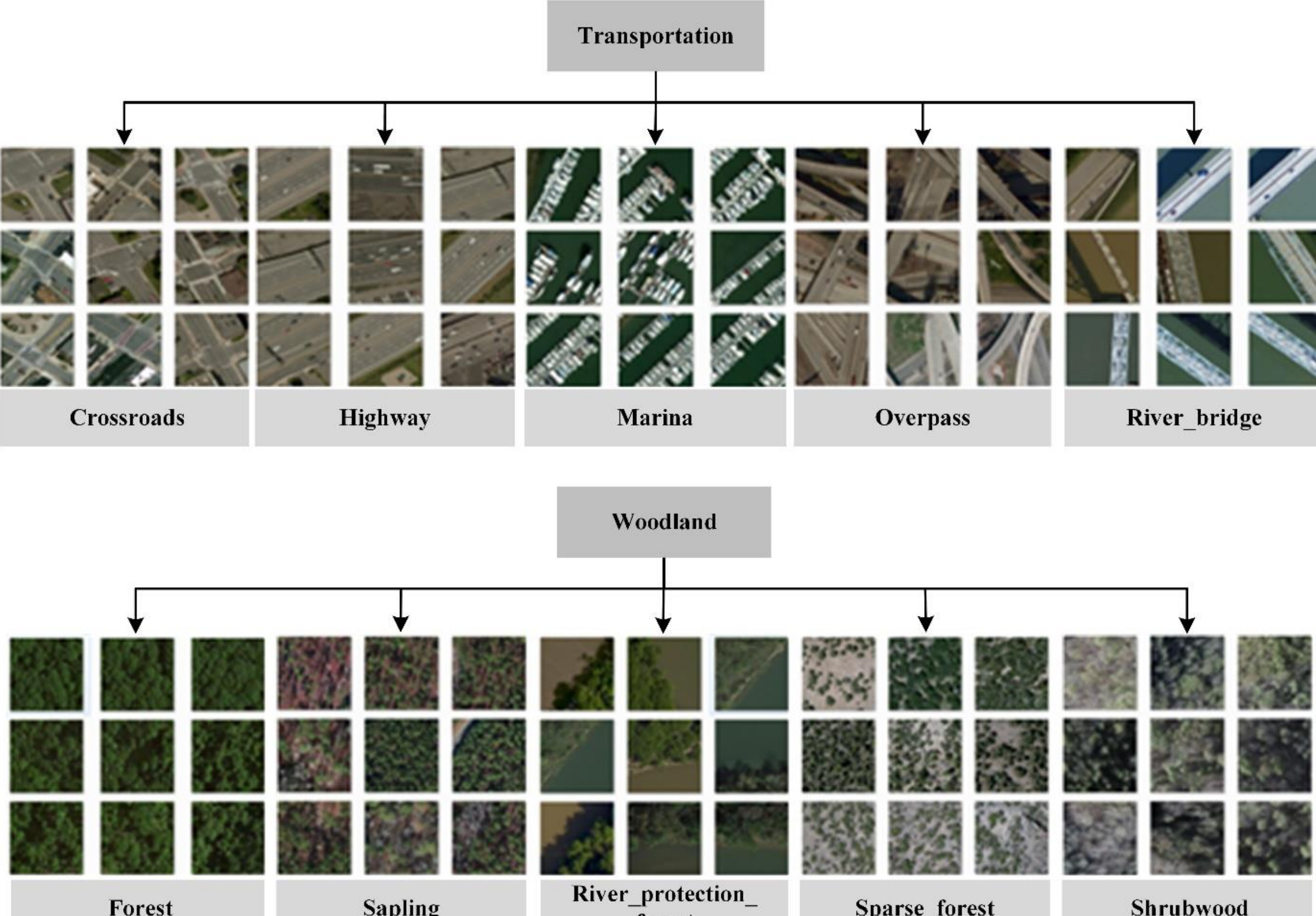

Figure 8: Root and leaf nodes of RSI-CB. Large category of transport corresponding to crossroads, highway, marinas, overpass, river_bridge, and other categories. The next figure is a large category of woodland, corresponding to the forest, sapling, river_protection_forest, sparse_forest, shrub wood, and other categories.

### 4.3 Distribution Characteristics of RSI-CB

We finally build six large categories, namely, agricultural land, construction and facilities, transportation and facilities, water and water conservancy facilities, woodland, and other land category according to Sections 3.1 and 4.2. The RSI-CB128 benchmark has 45 subcategories of approximately 36,000 images, with an average of approximately 800 images per class. The RSI-CB256 benchmark has 35 subcategories of approximately 24,000 images with an average of approximately 690 images per class. The large categories correspond to its subclasses, as shown in Table 3. The unique objects shown in brackets presented in italics belong to RSI-CB128. The italicized "airplane" is a unique class of RSI-CB256. Other classes are common to both. The large categories of transportation and facilities, woodland, and water and water conservancy facilities contain more subcategories, and the distribution of each category is shown in Figure 9、Figure 10、Figure 11 are the samples of RSI-CB128 and RSI-CB256 benchmarks, respectively.

Table 3. The corresponding of large categories to sub-categories in RSI-CB128 and RSI-CB256

| Large Class | Subclass |
|---|---|
| **Cultivated land** | green_farmland、dry_farm、bare_land |
| **Woodland** | artificial_grassland 、sparse_forest、forest、mangrove、river_protection_forest、shrubwood、sapling、*(natural_grassland*、*city_green_tree*、*city_avenue)* |
| **Transportation and facility** | airport_runway、avenue、highway、marina、parkinglot、crossroads bridge、*(city_road*、*overpass*、*rail*、*fork_road*、*turning_circle*、*mountain_road）*、*airplane* |
| **Water area and facility** | coastline、dam、hirst、lakeshore、river、sea、stream |
| **Construction land and facility** | city_building、container、residents、storage_room、pipeline、town 、*(tower*、*grave)*、 |
| **Other land** | desert、snow_mountain、mountain 、sandbeach |

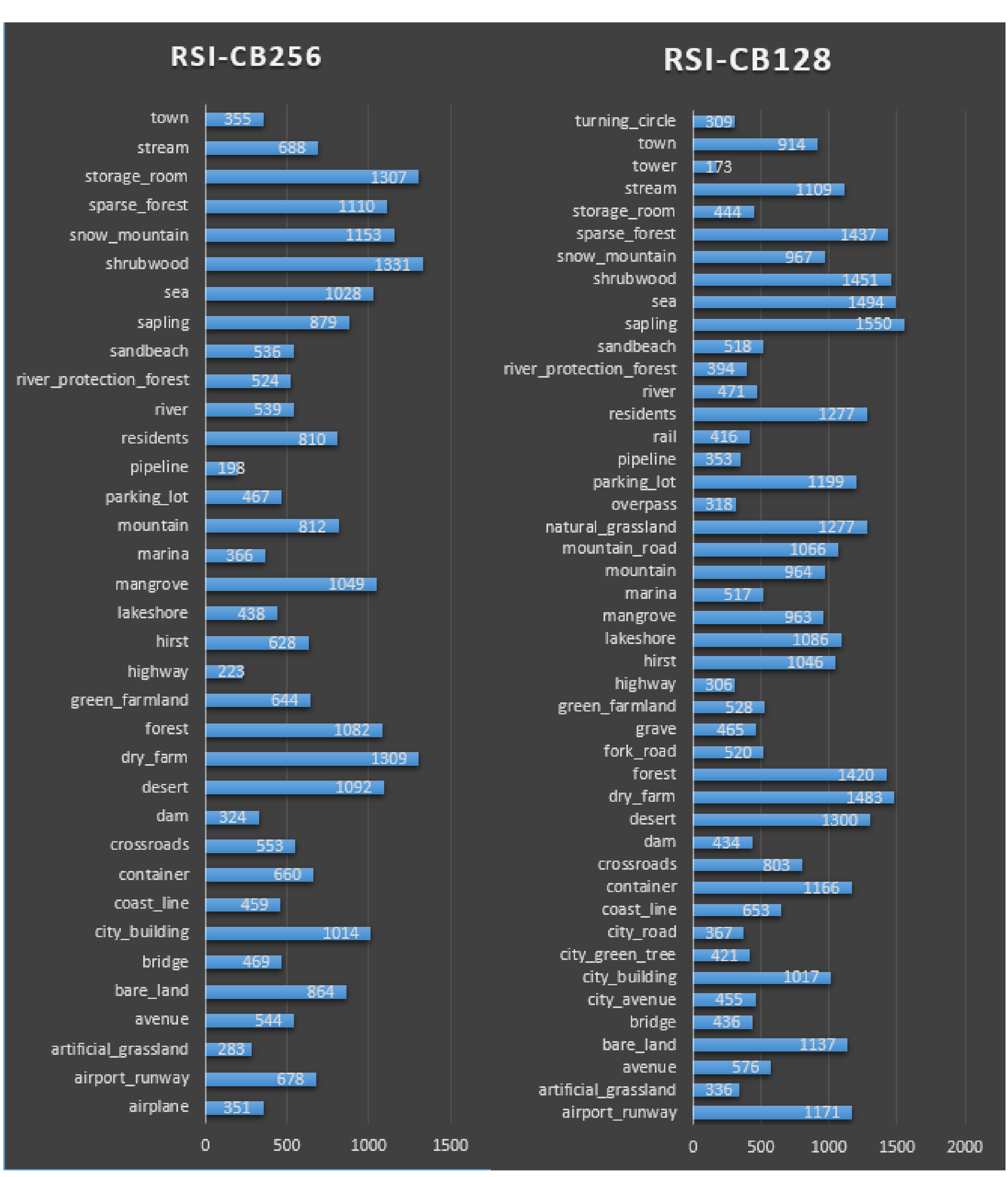

Figure 9: Number of distributions of RSI-CB256 and RSI-CB128 benchmarks for each category.

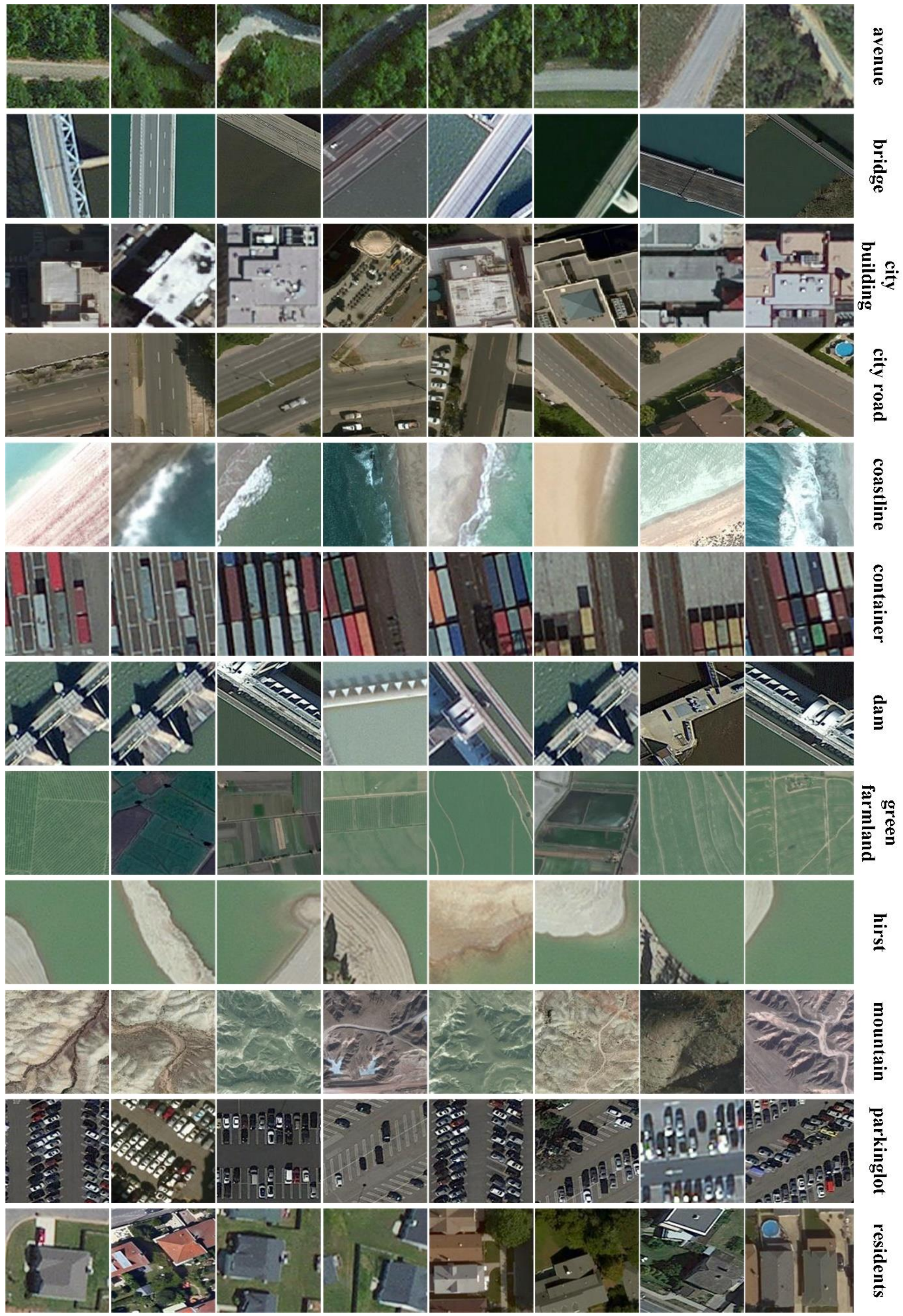

Figure 10: Samples of RSI-CB128

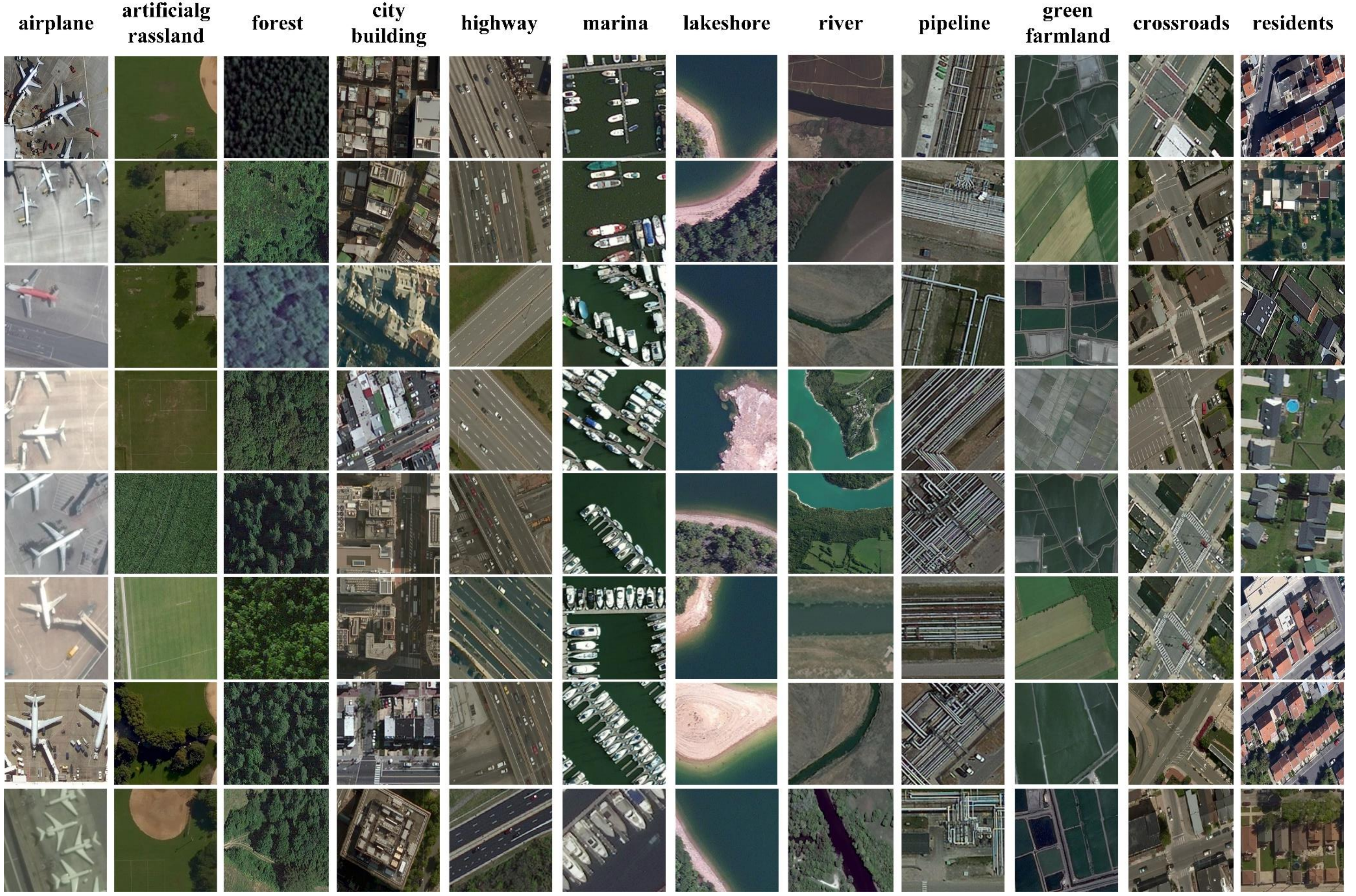

Figure 11: Samples of RSI-CB256

### 4.4 Comparison of RSI-CB with Some Remote Sensing Datasets

(1) Spatial resolution

RSI-CB has a higher spatial resolution than other existing datasets, which can be seen clearly from the image memory size. As such, AID is approximately 20–80 kb with an image size of 600 × 600; the NWPU-RESISC45 dataset is approximately 10–20kb with an image size of 256 × 256. For this study, each image is almost 193 kb for RSI-CB256 and almost 49 kb for RSI-CB128. Higher spatial resolution means higher information details, and the object's characteristic information is more comprehensive, which would be useful for object recognition.

(2) Scale of benchmark

RSI-CB is a larger-scale benchmark than any other databases which have approximately 60,000 images. RSI-CB128 has more than 36,000 images and 45 categories with 800 images per category. RSI-CB256 has more than 24,000 images and 35 categories with 690 images per category. **As for relatively large scale benchmark recently, there are 10,000 images, 30 categories and about 330 images per category for AID and 31,500images, 45 categories with each category about 700 images.**

(3) Construction model

Unlike the previous database construction model, RSI-CB makes full use of crowdsource geographic data, which has three contributions for remote sensing image dataset construction. First, we use the crowdsource for high-efficiency and high-quality method to build remote sensing benchmark as well as the possibility of continued expansion in terms of diversity, richness, and scale of benchmark. Second, we reflect the geographical significance of the geographic entity itself by constructing remote sensing dataset with crowdsource geographical data and making use of the criteria of land-use classification in China as well as in combination with ImageNet hierarchical grading mechanism. Finally, we can use the computer to achieve self-learning and learn the overall characteristics of the objects according to crowdsource massive data in future, which can aid in automatic labeling and recognition purposes to understand the image further.

(4) Diversity of benchmark

**Database diversity metrics can be divided into intra-class diversity and inter-class diversity. And diversity can be measured by similarity, for the same class, if the higher of similarity, said the lower the diversity, which is same to intra-class diversity. Therefore, we use the image similarity measure to verify the intensity of the image diversity, through**

the Bhattacharyya distances(Guorong et al., 1996) to calculate the difference between vectors to reflect degree of image similarity. In the statistical theory, Bhattacharyya distance is used to measure the similarity of two discrete or continuous probability distributions as an approximation of the overlap between the two statistical samples.

We randomly selected the sample categories of ‘river’ and ‘parking’ among four benchmarks with 100 images for each category based on the number limit for UC-Merced benchmark to calculate the inter-class similarity and intra-class similarity between the ‘river’ and the ‘parking’. We calculate the similarity between each of the image with any other image, then averaging the summed values Figure 13 shows the result of similarity measurement among four benchmarks. For inter-class similarity measurement, it shows the average similarity of ’river’ and ‘parking’ for which we calculate the similarity of two categories respectively, and then add the value to take the average. As Figure 13 shows, RSI-CB256 has lower inter-class similarity than the existing relatively large benchmark including AID and NWPU-RESISC45 and a litter higher than UC-Merced. However, RSI-CB256 get the lowest intra-class similarity compared to another benchmarks which just in the range of 0.45 and 0.51

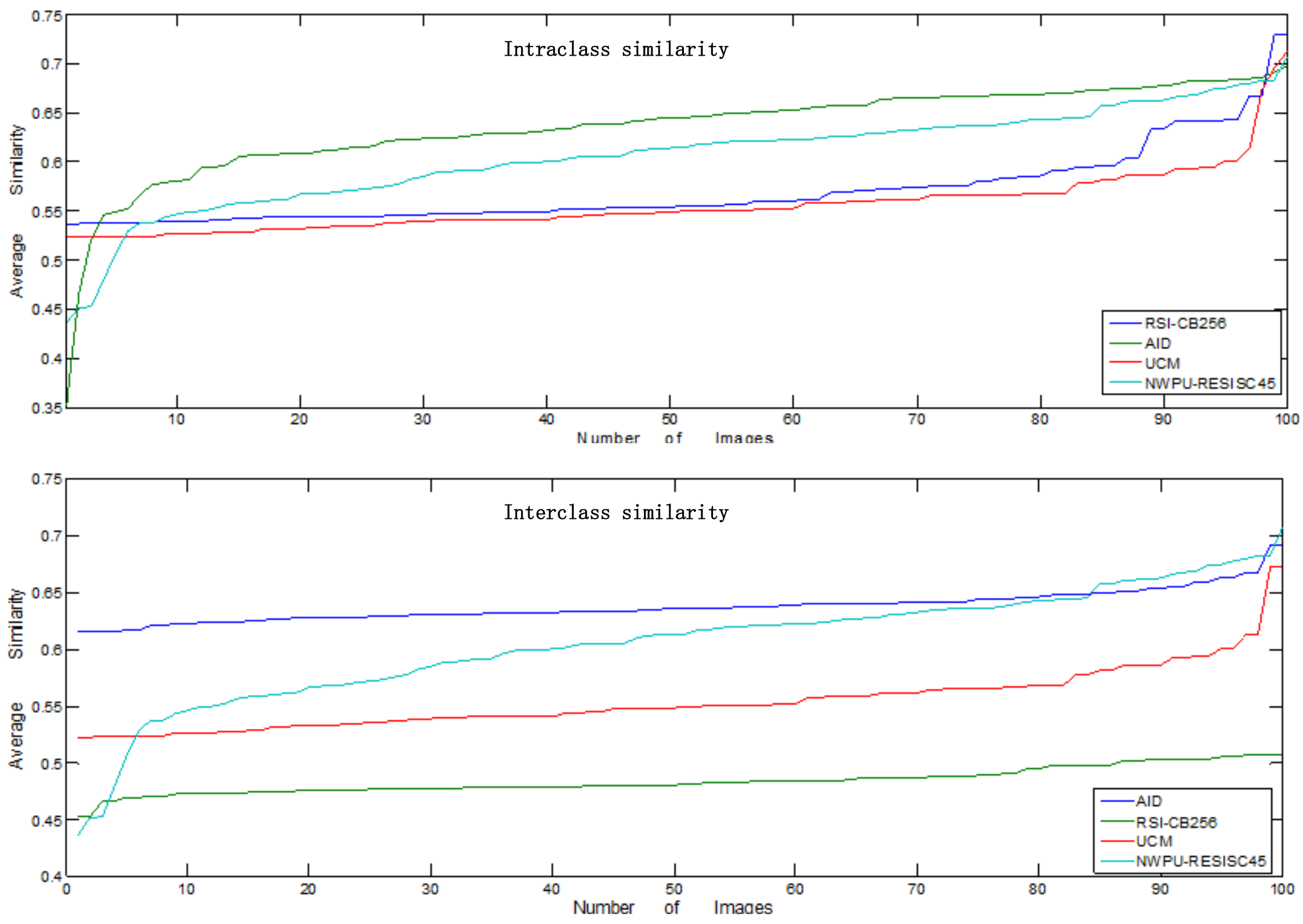

Figure 12. The measurement of intra-class similarity and inter-class similarity among four benchmarks.

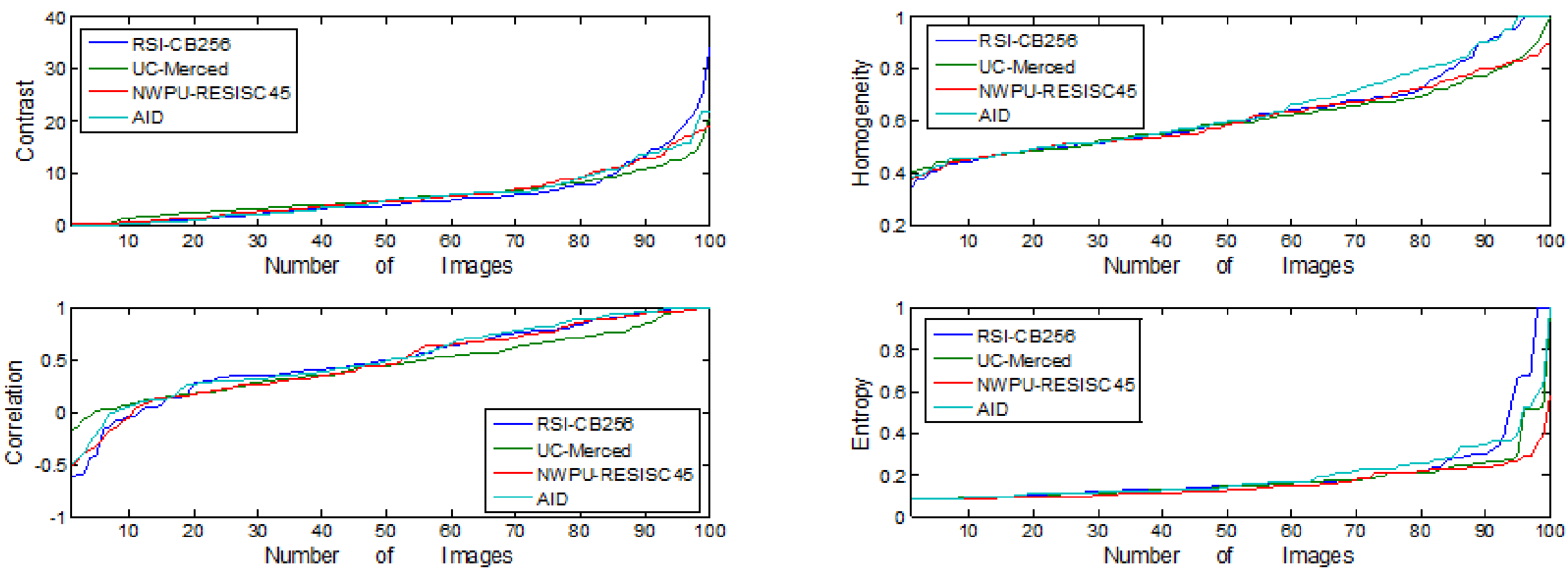

Figure 13. Compare image texture features among four benchmarks using GLCM.

**The co-occurrence matrix is defined by the joint probability density of the pixels at two positions, which not only reflect the distribution characteristics of the luminance but also the position distribution characteristic between the pixels with the same brightness or near brightness, which is the basis for defining a set of texture features. In order to be more intuitive to describe the texture of the co-occurrence matrix, we have described the following four parameters separately: a) Contrast: reflects the image sharpness and texture of the extent of the groove depth, the deeper texture groove means the greater contrast and the higher clarity; b) Homogeneity: measures the degree of local changes in image texture. Its larger value means the image texture of different regions lacking changes; c) Correlation: it reflects local gray image correlation. When the matrix pixel values are even equal, the correlation value is large; on the contrary, if the matrix pixel values vary widely, the correlation value is small; d) Entropy: it represents the complexity of the information within image.**

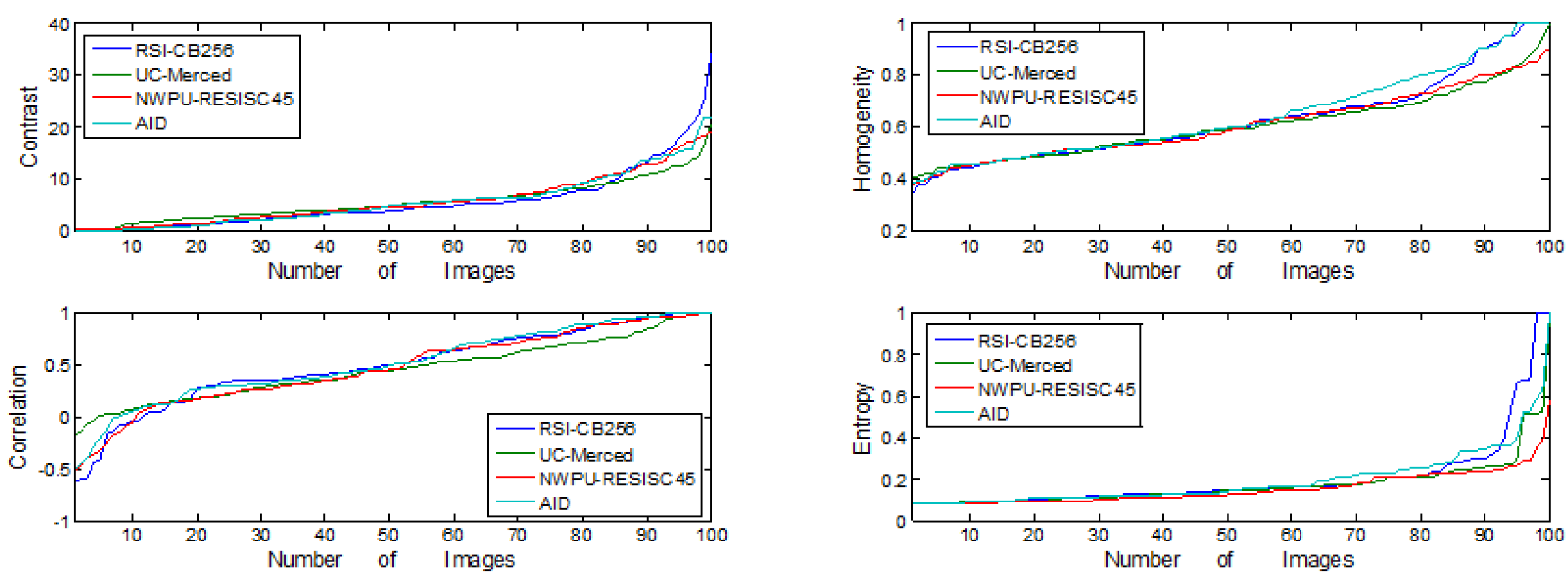

Figure 13 **shows the values of four parameters mentioned above among four benchmarks,**

**they are almost the same trend among them, however, the contrast, correlation and the entropy information of RSI-CB256 are relatively large, indicating that the RSI-CB256 image is clearer to recognize objects and features, and the high entropy suggests more complex of images which to some extent reflect the stronger diversity.**

Table 4 shows the comparison of several important factors for benchmark including the number of images and categories, spatial resolution, and size of images. SAT dataset has the largest number of images. SAT-4 has an average of more than 80,000, and SAT-6 has nearly 70,000 images per category, but the obvious drawback is that only six categories exist and the images are too small, which sacrifices the image size to obtain more images. **NWPU-RESISC45 has 45 categories as much as RSI-CB-128 and just a little less than us in scale. However, the difference between them is RSI-CB-128 most for object classification and NWPU-RESISC45 for scene classification because of different spatial resolution and image size. Besides that, the most important difference is that we construct RSI-CB with a higher efficient way which can update continuously when crowdsource data is new.**

Table 4: Comparison of RSI-CB with existing remote sensing datasets

| Database | Images | Categories | Average Per Category | Spatial Resolution(m) | Image Size |
|---|---|---|---|---|---|
| **UCM** | 2100 | 21 | 100 | 0.3 | 256*256 |
| **SAT-4** | 500,000 | 6 | 83333 | -- | 28*28 |
| **SAT-6** | 405,000 | 6 | 67500 | -- | 28*28 |
| **NLCD** | -- | 16 | -- | 30 | -- |
| **AID** | 10,000 | 30 | 333 | 0.5-8 | 600*600 |
| **NWPU-RESISC45** | 31,500 | 45 | 700 | 0.2-30 | 256*256 |
| **RSI-CB-128** | **36,707** | **45** | **800** | **0.22-3** | **128*128** |
| **RSI-CB256** | **24,747** | **35** | **690** | **0.22-3** | **256*256** |

Note: is the early image classification database, which contains three-year data of 1992, 2001, 2006, NLCD database records the pixels as categories, which is different from the recent database form.

In general, **RSI-CB have its advantages of spatial resolution, scale of quantity and novel way of constructing database compared some well-designed benchmark. However, RSI-CB** construction lies not only in the meaning of the database itself but also on the crowdsource data-based method for its potential application value **for weak-supervised learning to realize automatic marking and error correction of data.**

# 5 Experimental Analysis

## 5.1 Test Methods

(1) Model selection: handcrafted feature models versus DCNN (learning feature) models.

We use traditional handcrafted features and DCNNs based on end-to-end learning features as the test pipeline to test the performance of different methods on RSI-CB. As shown in Figure 14, for handcrafted features, we use SIFT(Lowe, 2004), CHI, LBP, and GIST as the description operators and then global mean pooling to construct eigenvectors for these methods. Finally, the SVM model is employed to classify the images. For end-to-end learning features, we use AlexNet (Krizhevsky et al., 2012), VGG-16 (Simonyan and Zisserman, 2014b), GoogLeNet (Szegedy et al., 2014), and ResNet (He et al., 2015a) to train RSI-CB.

(2) Training methods

We train DCNN models from scratch rather than fine tune them for the following reasons. First, considering the slight similarity between RSI-CB and ImageNet, the former is a satellite imagery and the latter is a natural imagery. The second reason is to define our own network structure. Third, the scale of RSI-CB is relatively large. **The last reason is convenient for other small-scale remote sensing image datasets to fine-tune.**

(3) Testing for model transfer performance

Transfer performance testing is based on the RSI-CB training model, which tests the model's ability to identify other databases. We select UC-Merced for the test database. We choose the lighter AlexNet-Conv3 model considering the small scale of the UC-Merced.

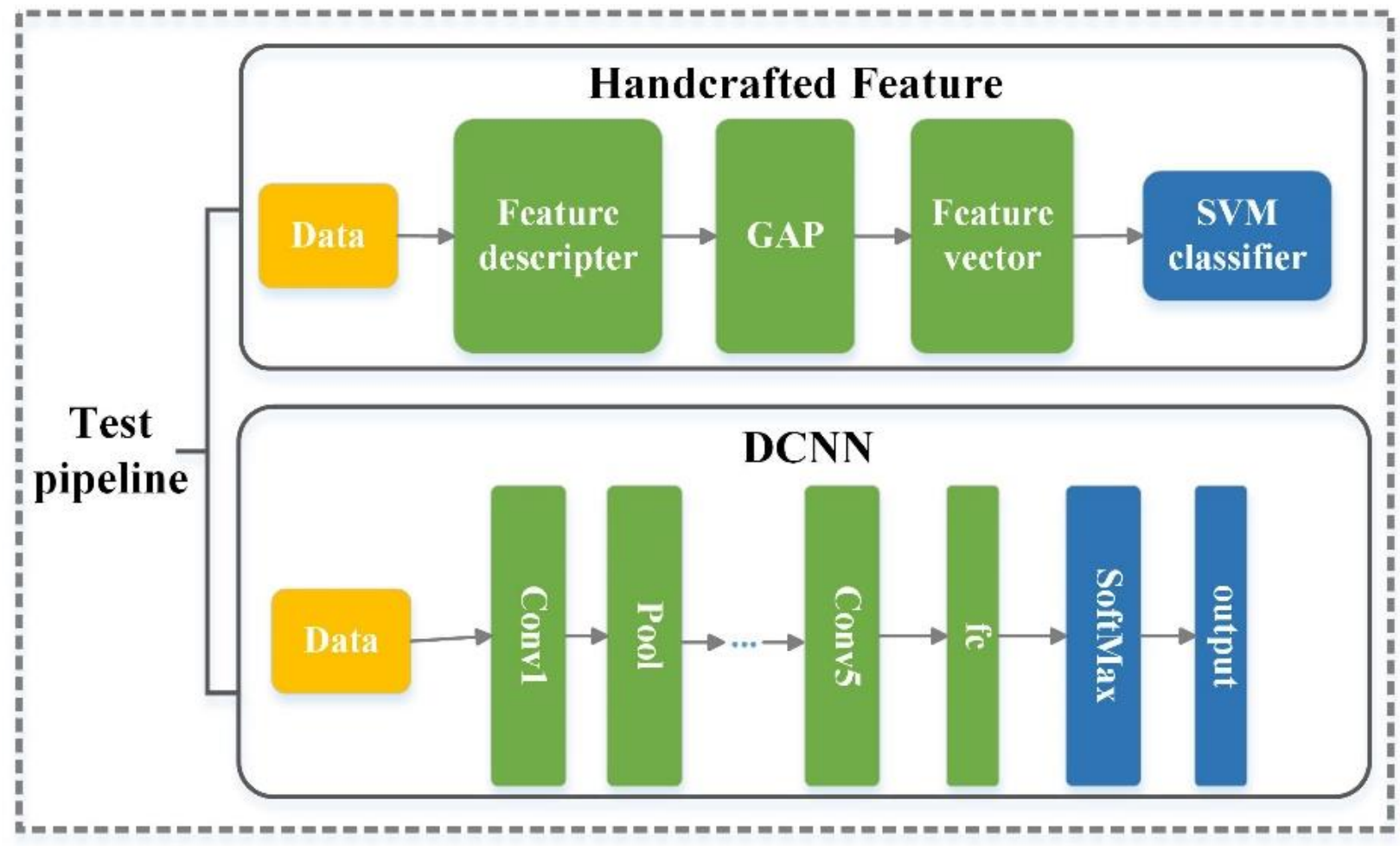

Figure 14: Test pipeline for RSI-CB

## 5.2 Data Organization

Data organization has three points, namely, selection of the training, validation, test sets for RSI-CB, data augmentation, and data organization for model transfer performance.

(1). Selecting data randomly. The training, validation, and test sets are randomly selected according to a certain proportion, and we disrupt labeling to further reflect the randomness of the data and objectivity.

(2). Data augmentation. We expand all the RSI-CB data for each image for cutting the fixed size from the middle point, upper-left, upper-right, lower-left, and lower-right corners of each image, and then flipping them before inputting to DCNNs, which have been expanded 10 times from the original data.

(3). Data organization for model transfer performance. We test whether the RSI-CB training model can transfer and the strength of the ability to transfer. We select 13 categories which are common for UC-Merced and RSI-CB256 as experimental data. The size of these two types of data is 256 × 256 because UC-Merced only has 100 images in each class. We select 1300 images for RSI-CB256 and UC-Merced as the test set.

### 5.3 Parameter Settings

(1). Handcrafted features

We refer to the method of (Xia et al., 2016) that uses four low-level features, namely, SIFT, LBP, CH, and GIST, as feature description operators. For the SIFT descriptor operator, we use a 16 × 16 fixed size grid with a step of eight pixels to extract all the descriptive operators in the gray image. Each dimension describing the operator uses the average pooling method to finally obtain the 128-dimensional image feature. For LBP, we use the usual eight directions to obtain the binary value and convert the 8-bit binary value to a decimal value for each pixel in the grayscale image. Finally, we obtain the LBP feature by calculating the frequency of 256 gray levels. For CH, we use RGB color space directly and quantize each channel into 32 bins. Therefore, we obtain the feature of images by connecting the statistical histogram in each channel. For GIST, we use the original parameter settings in (Oliva and Torralba, 2001) directly, using four scales and eight directions and 4 × 4 spatial grid for pooling. Finally, we obtain a 512 (4 × 8 × 4 × 4) dimension eigenvector.

(2). DCNN models

We retain most of the default parameters to train DCNN and fine-tune the learning rate and batch size. Our model can finally converge better although we make concessions on the computation time and the convergence rate. In addition, the vibration of loss function value is smaller, which is beneficial for improving the performance of our model. In the RSI-CB128 test, we implement a slight adjustment in the network because of the input size constraints of the network and adjust the pad for the convolution and pooling layers. We do not warp images because it will affect the real information of the images to some extent.

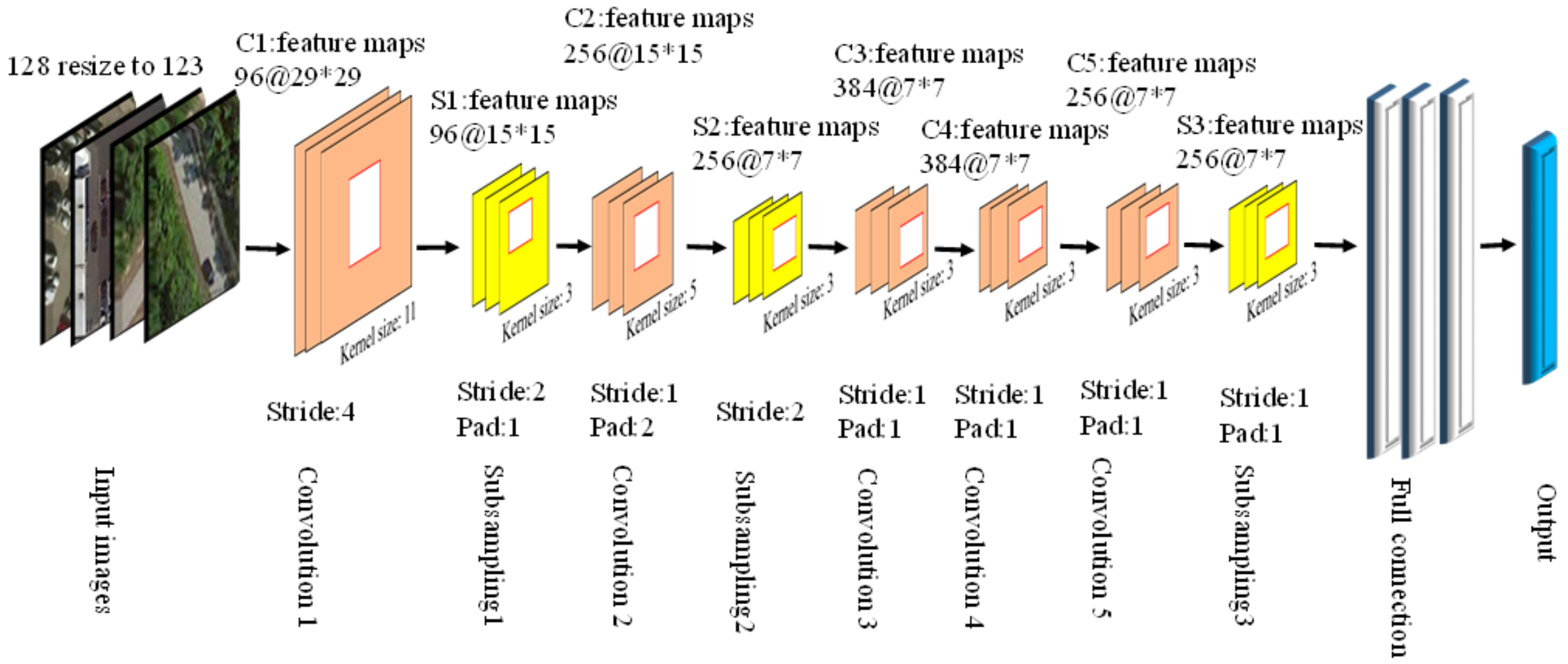

Figure 15: AlexNet model parameter settings in RSI-CB128.

### 5.4 Evaluation Methods

We use two common methods as the evaluation index of the training model, namely, overall accuracy (OA) and confusion matrix (Congalton, 1991).

(1). OA refers to the ratio of the number of categories that are correctly classified to the total number of categories. The OA value can be a good characterization of the overall classification accuracy. However, when categories are extremely imbalanced, the OA value is greatly affected by categories with more images.

(2). The confusion matrix can visually reflect the classification accuracy of each type of object. We can clearly determine the correct and wrong classification of each category in each row.

### 5.5 Experimental Results

The experimental results are divided into three parts, namely, classification results based on handcrafted features and DCNN and model transfer ability test results. We use different test methods in the first two parts of the results to evaluate our benchmark and differentiate the tests in the UC-Merced, SAT-6, and SAT-4 datasets to compare different performances. For the transfer ability test, we extract 13 common categories of RSI-CB256 and UC-Merced with 100 images per category and name them RSI-CB256-13 and UC-Merced-13, respectively. Then, we train AlexNet-Conv3 with the remaining images in RSI-CB256-13. Finally, we validate the classification ability of the model on UC-Merced-13.

#### 5.5.1 Classification Results Based on Handcrafted Features

We use the UC-Merced and RSI-CB256 benchmarks with a uniform size of $256 \times 256$ as experimental data to ensure the fairness of results. We test the data 10 times and take their mean and standard deviation as the results.

Table5.OA test on RSI-CB256 and UC-Merced

| Methods | Our dataset(50%) | Our dataset(80%) | UC-Merced(50%) | UC-Merced(80%) |
|---|---|---|---|---|
| **SIFT** | 37.96±0.27 | 40.12±0.34 | 29.45±1.08 | 32.10±1.95 |
| **LBP** | 69.10±0.20 | 71.98±0.36 | 35.04±1.08 | 36.29±1.90 |
| **CH** | 84.08±0.26 | 84.72±0.33 | 42.79±1.06 | 46.21±1.05 |
| **Gist** | 61.74±0.35 | 63.59±0.45 | 45.38±0.70 | 46.90±1.76 |

Table 5 shows the mean and standard deviation of the OA based on the handcrafted features on the UC-Merced and RSI-CB256 benchmarks. The values in brackets refer to the ratio of the training set to the total number of datasets. The overall test results using the SIFT method for RSI-CB256 are relatively low, indicating that SIFT is inadequate to represent the characteristics of our remote sensing images. The best results are achieved using the CH method on RSI-CB256 with an accuracy of more than 80%. However, its result is not the best on UC-Merced because of the following two reasons. We have a clear advantage in the number of images and some differences between RSI-CB256 and UC-Merced exist, which are mainly reflected in the diversity and complexity of the database.

In addition, the experimental results show that no universal recognition algorithm exists for different datasets, and test results improve significantly when the training set increases, which indicates that the amount of data is still a key factor in improving the accuracy of recognition.

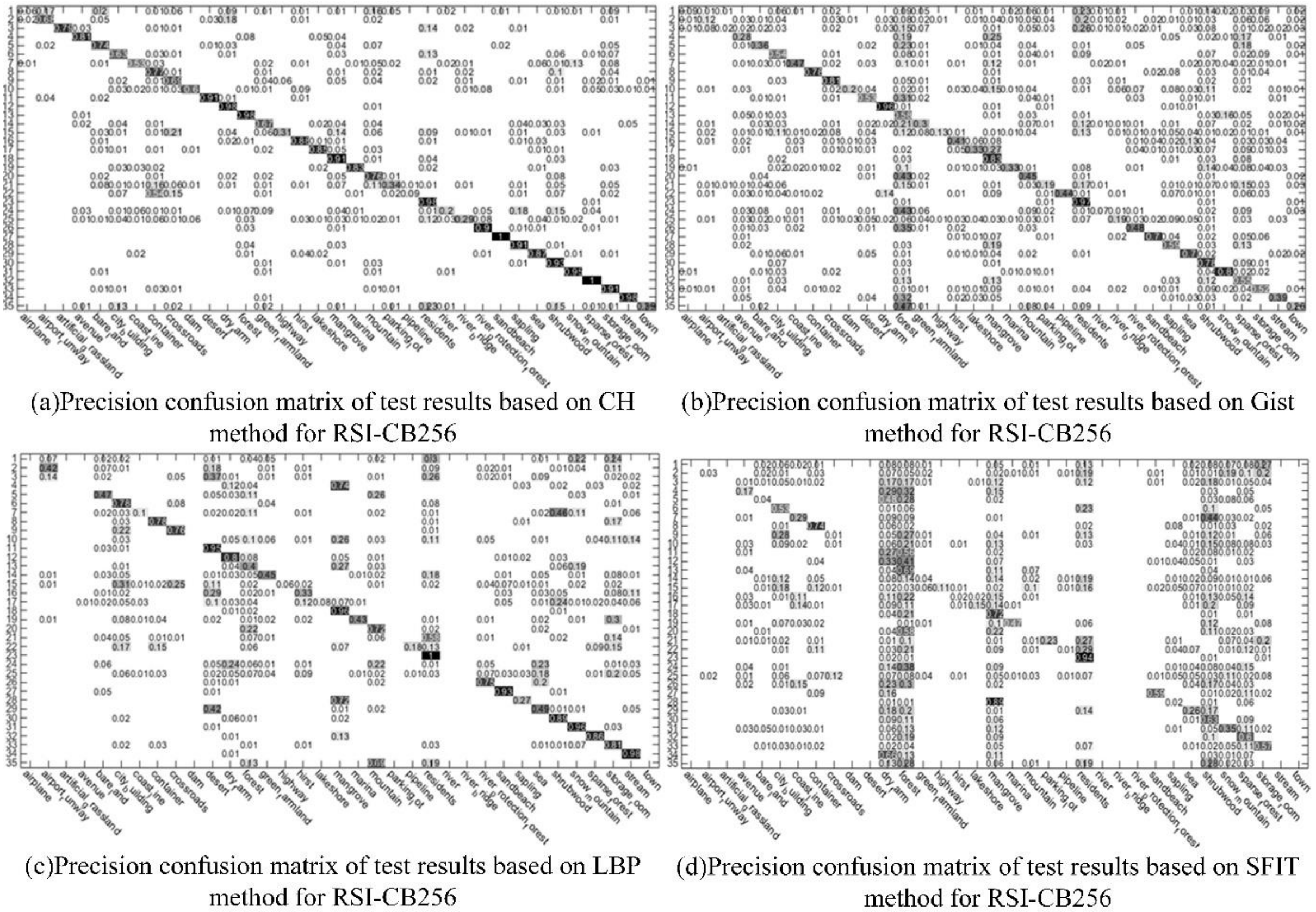

(a)Precision confusion matrix of test results based on CH method for RSI-CB256

(b)Precision confusion matrix of test results based on Gist method for RSI-CB256

(c)Precision confusion matrix of test results based on LBP method for RSI-CB256

(d)Precision confusion matrix of test results based on SFIT method for RSI-CB256

Figure 16: Precision confusion matrix of test results based on handcrafted features for RSI-CB256.

Figure 16 shows the precision confusion matrix of test results using handcrafted features on the RSI-CB256 benchmark. From the precision confusion matrix above, the classification precision of a single category differs in various methods, and each category of its classification performance is consistent with OA. SIFT is still inadequate for classifying most categories. The classification accuracy of nearly half of the categories based on CH is more than 0.8, and LBP also achieves good results. However, these methods perform poorly for some categories (e.g., airplane and parking lot) because of the small difference between category features and the large overlapping ratio of the interclass feature distribution. Furthermore, richer features of categories exist. Consequently, handcrafted methods are usually used for describing low-level features and cannot fully describe the distribution of features.

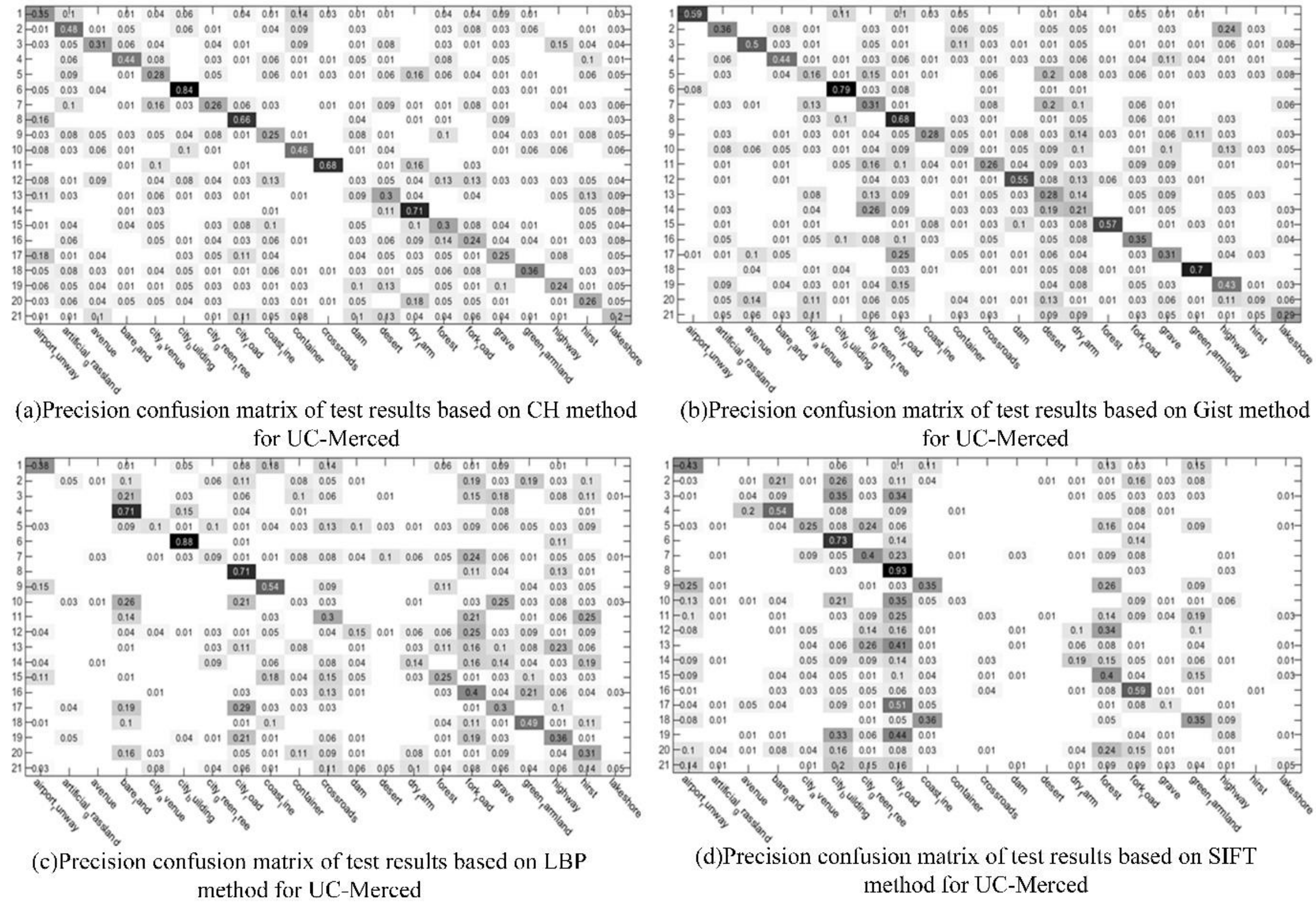

(a)Precision confusion matrix of test results based on CH method for UC-Merced

(b)Precision confusion matrix of test results based on Gist method for UC-Merced

(c)Precision confusion matrix of test results based on LBP method for UC-Merced

(d)Precision confusion matrix of test results based on SIFT method for UC-Merced

Figure 17: Precision confusion matrix of test results based on the handcrafted features for UC-Merced.

The performance based on handcrafted features on UC-Merced shown in Figure 17 is much worse than the test results in RSI-CB256, where only few categories reach 0.8 precision. The main reason lies in the obvious scale advantage of the RSI-CB256 benchmark. In addition, RSI-CB256 is an object-centered benchmark for most categories, and UC-Merced is based on complex scenes for most categories. Each image contains more complex information and robust visual features.

To sum up, the method based on color, texture, and other handcrafted features cannot represent complex object feature recognition because manually designing low-level visual features and obtaining the optimal eigenvalue and expressing it appropriately directly affect the accuracy of classification recognition.

### 5.5.2 Classification Results Based on DCNN

We test RSI-CB and compare it with other remote sensing datasets based on the DCNN method. We use the AlexNet-Conv3 network with only three convolutions to test SAT-6 and SAT-4 because of their small size (28 × 28).

Tabel6.OA of training and test on datasets using DCNN

| **Methods** | **RSI-CB256** | **RSI-CB128** | **UCM-Merced** | **SAT6** |
|---|---|---|---|---|

| **AlexNet-3ConV** | - | 96.68%/85.59% | 78.69%/70.00% | 98+%/97.53% |
|---|---|---|---|---|
| **AlexNet** | 96+%/94.78% | - | 74.82/65.53% | - |
| **VGG-16** | 98+%/95.13% | 98+%/94.39% | - | - |
| **GoogLeNet** | 98+%/94.07% | 96.21%/91.91% | 76.58%/67.62% | - |
| **ResNet** | 98+%/95.02% | 96.81%/93.56% | - | - |

Table 6 describes the overall performance of different models based on the DCNN for different datasets. The test results of the models for RSI-CB256 are more than 90%. The classification accuracy of GoogLeNet results is lower than that of the network structure of GoogLeNet because the wide range of macro performance of objects for RSI-CB does not require high-precision features to represent, which may be slightly different from the natural image. The test results of accuracy on RSI-CB128 benchmark show that VGGNet and ResNet have been significantly higher than other models, indicating that it is beneficial for improving the overall recognition performance of models by keeping the model deeper when other conditions are constant. The recognition accuracy of GoogLeNet in RSI-CB128 and RSI-CB256 is less than AlexNet, which contradicts the findings mentioned above. The overall recognition performance of the models does not rely only on the network depth, but also depends on different network structures and the data itself. VGGNet performs unexpectedly better than the top-1 of ResNet for natural image recognition, which further shows that the performance of models is undoubtedly related to the benchmark itself and the difference between remote sensing and natural image databases. The results of our benchmark are greatly improved due to the obvious advantage of data scale compared with the UC-Merced. The SAT-6 results are highly satisfactory. This is probably because of the small number of categories and low complexity of images with many similar features, which can easily obtain excellent performance.

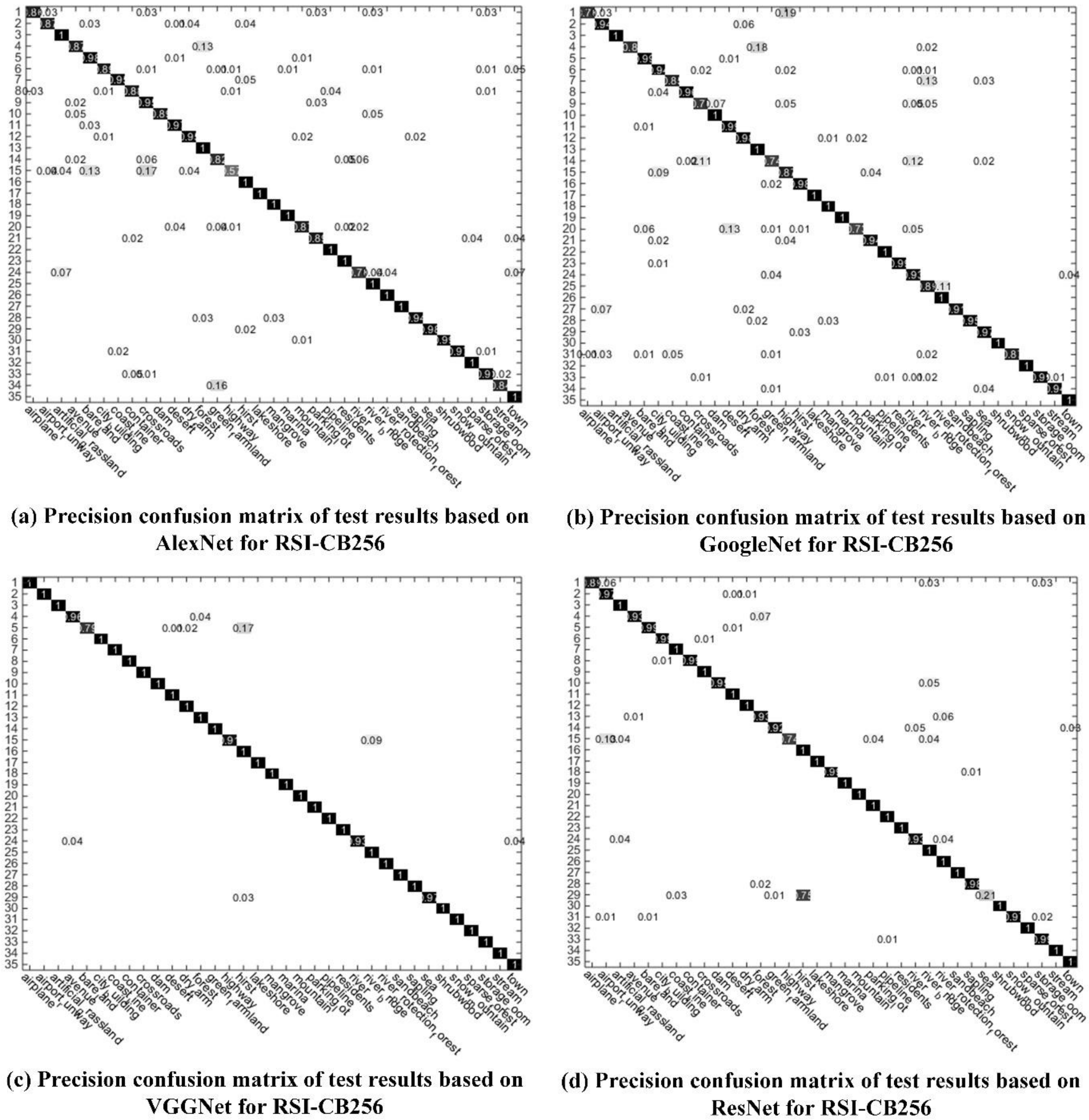

**(a) Precision confusion matrix of test results based on AlexNet for RSI-CB256**

**(b) Precision confusion matrix of test results based on GoogleNet for RSI-CB256**

**(c) Precision confusion matrix of test results based on VGGNet for RSI-CB256**

**(d) Precision confusion matrix of test results based on ResNet for RSI-CB256**

Figure 18: Precision confusion matrix of test results based on DCNN models for RSI-CB256.

Figure 18 reflects the test results based on the DCNN methods for the RSI-CB256 benchmark. Only a few categories of classification accuracy have less than 0.8 precision, and the precision of nearly half of the categories is 0.9 or more. DCNN shows a stronger recognition and non-linear fitting abilities than handcrafted methods. In addition, it benefits from its database scale. Thus, DCNN can learn features from different categories. The classification performance of each category is similar in different networks, but few categories show unusual classification precision, which may be attributed to the random initialization and different structures of models. VGGNet and ResNet have achieved higher precision in each category than AlexNet and GoogLeNet, which further indicates that deeper networks are beneficial for improving model identification performance.

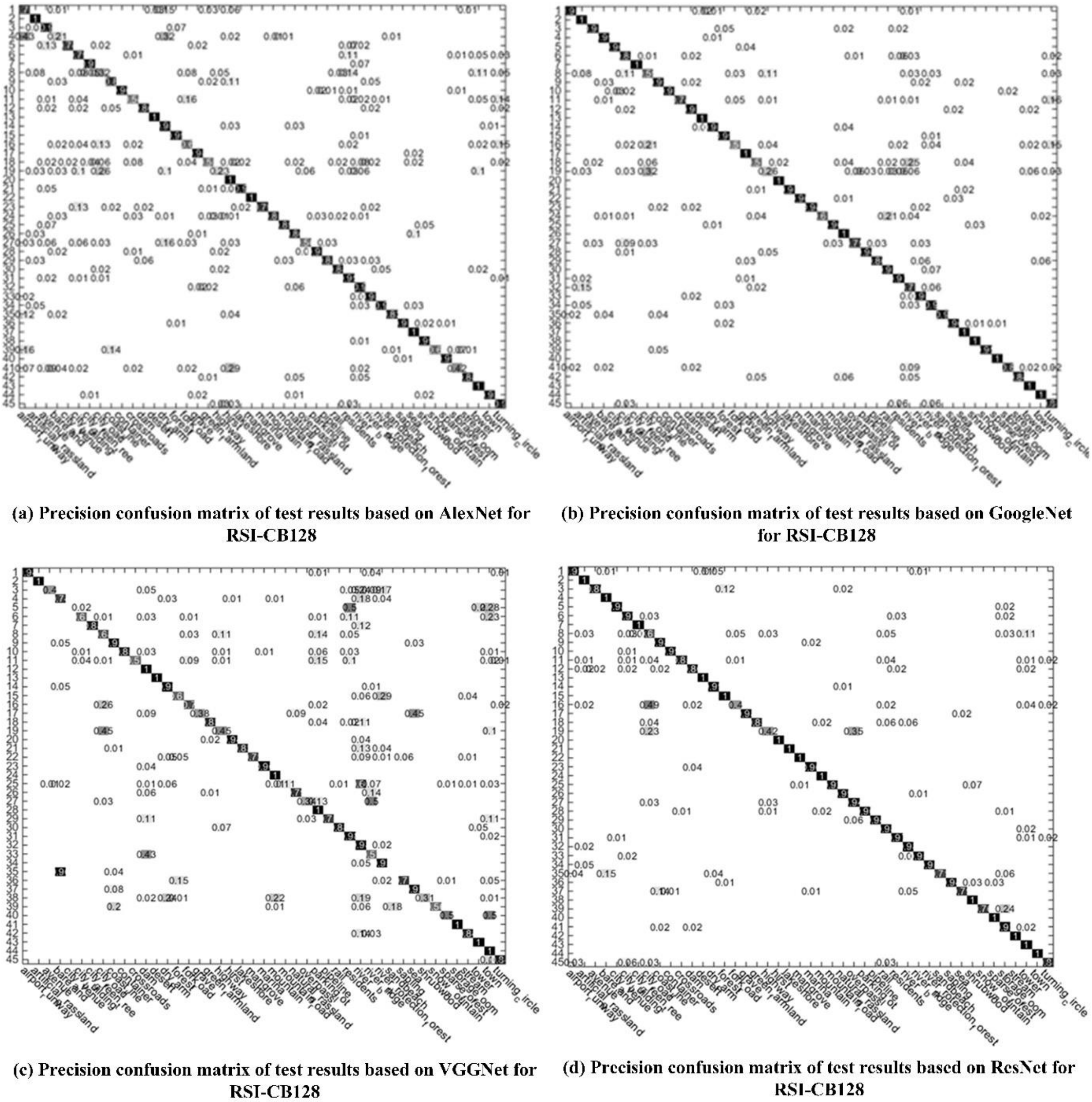

(a) Precision confusion matrix of test results based on AlexNet for RSI-CB128

(b) Precision confusion matrix of test results based on GoogleNet for RSI-CB128

(c) Precision confusion matrix of test results based on VGGNet for RSI-CB128

(d) Precision confusion matrix of test results based on ResNet for RSI-CB128

Figure 19: Precision confusion matrix of test results based on DCNN models for RSI-CB128.

Figure 19 shows the precision confusion matrix of the four network test results trained on the RSI-CB128 benchmark. The four network models can identify most of the features as well as RSI-CB256. However, several classes in several models are not ideal for the recognition results, such as highway and folk road, because the difference between subclasses is too small in the big category of transportation and facilities, resulting in large overlapping ratios of feature space between classes, which causes further confusion when models attempt to classify these subclasses. Therefore, this experiment also verifies the influence of inter-class variance in image classification accuracy.

### 5.5.3 Evaluation of Model Transfer Capability

According to the assumption that samples of hypothesis and real spaces conform to the same distribution, the representation of samples is the key point of dataset quality. A general database evaluation index often evaluates a dataset and considers the classification precision of the test set performance of models trained using a training set. The distribution of spatial features in real data and of test samples are different. To improve model performance, we should allow the model to learn more data that are closer to the real distribution.

We select 13 common categories in UC-Merced and RSI-CB256 as test sets, with each category containing 100 images according to the scale of UC-Merced, namely, UC-Merced-13 and RSI-CB256-13, to verify the transfer ability of the model trained on our dataset. We use the remaining images of RSI-CB256 for the 13 categories as training set and select AlexNet-Conv3 as the training model considering the small-scale samples, and the training models are tested in UC-Merced-13 and RSI-CB256-13.

Table 7. Test results based on AlexNet-Conv3 for RSI-CB256-13 and UC-Merced-13

| Dataset | Train | Test |
|---|---|---|
| RSI-CB256-13 | 90%+ | 86.32% |
| UC-Merced-13 | -- | 74.13% |

Table 7 shows the results of the test on RSI-CB-13 and UC-Merced-13. The two datasets have differences in visualization, but they are strongly consistent in the overall test results. As shown above, the model trained on our dataset still has good transfer ability for UC-Merced whose classification precision is 74.13%, which is approximately 12 percentage points less than RSI-CB, indicating that our data samples can represent real-world samples better. **However, the result can also provide evidence that although there are same categories and its quantity between two benchmarks which are the same type of images, the accuracy of UC-Merced-13 is decreased. The result maybe can explain no matter how similar between two benchmarks, there are always exit feature gap between them which for the lager benchmark is not enough to learn all the features to train a learning model for the task of small benchmark classification and that can also explain the importance of scale for benchmark to fit the overall data characteristics as much as possible.**

## 6 Conclusions and Future Works

Crowdsource data have become the research focus of international geographic information science because of several remarkable features, such as real time classification, fast spread speed, robust information, low cost, and massive data. The RSI-CB based on crowdsource data provides new ideas and research directions for the establishment and improvement of remote sensing datasets. The RSI-CB has six categories that are based on the land-use classification standard in China, and each category has several subcategories, which have significantly improved the number of categories and images compared with other remote sensing datasets. The classification experiment conducted in several traditional deep learning networks shows that the classification accuracy of RSI-CB is higher than that of other datasets because of its larger and higher spatial resolution.

RSI-CB can continue to expand in category, quantity of objects, and diversity based on crowdsource data, and these kinds of data are updated rapidly and increase massively in number. However, artificial work is required to obtain the appropriate categories and images given the complexity of the remote sensing image itself and the spatial resolution, which limit the construction speed to some extent, resulting in a large amount of unavailable POI data.

At present, the existing geographical data labels are manually marked, which requires a huge human, material and financial resources. In addition, annotations come from different people leading to the quality of inconsistencies and multiple checks which would increase the workload. If only use a small number of labeled samples, it is often difficult for a learning model to have strong generalization trained by them. On the other hand, if you use only a small number of "expensive" tagged examples without using a large number of "cheap" unlabeled examples, it is a great waste of data resources. In practice, unlabeled examples can be used to help improve learning performance as long as the link between the unmarked sample distribution and the learning model can be reasonably established. Therefore, if combine RSI-CB which has been correctly marked with unlabeled or incomplete data to learn together, we can not only greatly increase the amount of data to provide the generalization of the model performance, in addition, we can also correct errors in manual markers or to achieve the correct markings of unmarked data based on RSI-CB and weak-supervised learning.

The traditional remote sensing image object recognition usually includes visual interpretation and computer interpretation. The defects of visual interpretation are time consuming, consumable and labor, and the accuracy of computer interpretation needs to be further improved. Inspired by the success of using DCNN on natural image classification, remote sensing experts introduced DCNN to remote sensing image classification and other recognition tasks which is the current frontier and highlight of remote sensing image processing. However, the remote sensing image field is lacking a large-scale benchmark. It is very important to efficiently build a large-scale multi-category remote sensing image benchmark. RSI-CB is based on the Chinese land use classification standard, which contains six categories of agricultural land, construction land and facilities, water and water conservancy facilities, woodland, and other lands.

Experiments show that RSI-CB benchmark use different depth models to achieve good image recognition accuracy and have higher classification accuracy when transferring to other remote sensing image databases, which further proves that the RSI-CB classification benchmark plays an important role in image object recognition. Therefore, we can reasonably believe that RSI-CB can be applied to different types land recognition of remote sensing images, which can service in the relevant departments, making land use more reasonable.

For the current remote sensing image database, such as UCM, AID achieved good results in scene classification research. We believe that the scalable, high-quality and diversity, large-scale, and multi-category RSI-CB will become a new and important remote sensing image benchmark to develop new ideas because of its application value or its method of construction.

## Acknowledgments:

This research has also been supported by National Natural Science Foundation of China (41671357, 41571397, and 41501442) and Natural Science Foundation of Hunan Province (2016JJ3144).

This research has also been supported in part by the Open Research Fund Program of Shenzhen Key Laboratory of Spatial Smart Sensing and Services (Shenzhen University), and the Scientific Research Foundation for the Returned Overseas Chinese Scholars, State Education Ministry (50-20150618).

## Reference

Basu, S., Ganguly, S., Mukhopadhyay, S., Dibiano, R., Karki, M., Nemani, R., 2015. DeepSat: a learning framework for satellite imagery. Computer Science, 1-10.

Basu, S., Ganguly, S., Nemani, R.R., Mukhopadhyay, S., Milesi, C., Votava, P., Michaelis, A., Zhang, G., Cook, B.D., Saatchi, S.S., 2014. A Semi-Automated Machine Learning Algorithm for Tree Cover Delineation from 1-m Naip Imagery Using a High Performance Computing Architecture, AGU Fall Meeting.

Chen Baiming, Z.X., 2007. National Standards Interpretation of "Land Use Classification Status. Natural Resources 22, 994-1003.

Chen, Y., Lin, Z., Zhao, X., Wang, G., Gu, Y., 2014. Deep Learning-Based Classification of Hyperspectral Data. IEEE Journal of Selected Topics in Applied Earth Observations & Remote Sensing 7, 2094-2107.

Cheng, G., Han, J., Guo, L., Liu, Z., Bu, S., Ren, J., 2015. Effective and Efficient Midlevel Visual Elements-Oriented Land-Use Classification Using VHR Remote Sensing Images. IEEE Transactions on Geoscience and Remote Sensing 53, 4238-4249.

Cheng, G., Han, J., Lu, X., 2017. Remote Sensing Image Scene Classification: Benchmark and State of the Art. Proceedings of the IEEE PP, 1-19.
Cheng, G., Han, J., Zhou, P., Guo, L., 2014. Multi-class geospatial object detection and geographic image classification based on collection of part detectors. Isprs Journal of Photogrammetry and Remote Sensing 98, 119-132.
Cheng, G., Zhou, P., Han, J., 2016. Learning Rotation-Invariant Convolutional Neural Networks for Object Detection in VHR Optical Remote Sensing Images. IEEE Transactions on Geoscience & Remote Sensing 54, 7405-7415.
Congalton, R.G., 1991. A review of assessing the accuracy of classification of remotely sensed data. Remote Sensing of Environment 37, 35-46.
Cui, S., 2016. Comparison of approximation methods to Kullback–Leibler divergence between Gaussian mixture models for satellite image retrieval. Remote Sensing Letters 7, 651-660.
Deng, J., Dong, W., Socher, R., Li, L., Li, K., Feifei, L., 2009. ImageNet: A large-scale hierarchical image database, computer vision and pattern recognition, pp. 248-255.
Donahue, J., Hendricks, L.A., Guadarrama, S., Rohrbach, M., Venugopalan, S., Darrell, T., Saenko, K., 2015. Long-term recurrent convolutional networks for visual recognition and description. Elsevier.
Fry, J.A., Xian, G.S., Jin, S., Dewitz, J.A., Homer, C.G., Yang, L., Barnes, C.A., Herold, N.D., Wickham, J.D., 2011. Completion of the 2006 National Land Cover Database for the Conterminous United States. Photogrammetric Engineering & Remote Sensing 77, 858-864.
Girshick, R., 2015. Fast R-CNN. Computer Science.
Girshick, R., Donahue, J., Darrell, T., Malik, J., 2014. Rich Feature Hierarchies for Accurate Object Detection and Semantic Segmentation. 580-587.
Guorong, X., Peiqi, C., Minhui, W., 1996. Bhattacharyya distance feature selection, international conference on pattern recognition, pp. 195-199.
Haklay, M., 2010. How good is volunteered geographical information? A comparative study of OpenStreetMap and Ordnance Survey datasets. Environment and Planning B: Planning and Design 93, 3-11.
Haklay, M., Weber, P., 2008. OpenStreetMap: User-Generated Street Maps. IEEE Pervasive Computing 7, 12-18.
Han, J., Zhang, D., Cheng, G., Guo, L., Ren, J., 2015. Object Detection in Optical Remote Sensing Images Based on Weakly Supervised Learning and High-Level Feature Learning. IEEE Transactions on Geoscience & Remote Sensing 53, 3325-3337.
He, K., Zhang, X., Ren, S., Sun, J., 2015a. Deep Residual Learning for Image Recognition, Computer Vision and Pattern Recognition, pp. 770-778.
He, K., Zhang, X., Ren, S., Sun, J., 2015b. Delving Deep into Rectifiers: Surpassing Human-Level Performance on ImageNet Classification. 1026-1034.
Heipke, C., 2010. Crowdsourcing geospatial data. Isprs Journal of Photogrammetry & Remote Sensing 65, 550-557.
Homer, C., Dewitz, J., Fry, J., Coan, M., Hossain, N., Larson, C., Mckerrow, A., Vandriel, J.N., Wickham, J., 2007. Completion of the 2001 National Land Cover Database for the Conterminous United States. Photogrammetric Engineering & Remote Sensing 77, 858-864.

Hu, F., Xia, G., Hu, J., Zhang, L., 2015. Transferring Deep Convolutional Neural Networks for the Scene Classification of High-Resolution Remote Sensing Imagery. Remote Sensing 7, 14680-14707.

Ji, S., Xu, W., Yang, M., Yu, K., 2013. 3D Convolutional Neural Networks for Human Action Recognition. IEEE Transactions on Pattern Analysis & Machine Intelligence 35, 221.

Krizhevsky, A., Sutskever, I., Hinton, G.E., 2012. ImageNet classification with deep convolutional neural networks, neural information processing systems, pp. 1097-1105.

Levin, G., Newbury, D., McDonald, K., Alvarado, I., Tiwari, A., and Zaheer, M., 2016. Terrapattern: Open-Ended, Visual Query-By-Example for Satellite Imagery using Deep Learning.

Liu, W., Anguelov, D., Erhan, D., Szegedy, C., Reed, S., Fu, C.Y., Berg, A.C., 2016a. SSD: Single Shot MultiBox Detector.

Liu, Y., Zhang, Y., Zhang, X., Liu, C., 2016b. Adaptive spatial pooling for image classification. Pattern Recognition 55, 58-67.

Lowe, D.G., 2004. Distinctive Image Features from Scale-Invariant Keypoints. International Journal of Computer Vision 60, 91-110.

Melgani, F., Bruzzone, L., 2004. Classification of hyperspectral remote sensing images with support vector machines. IEEE Transactions on Geoscience and Remote Sensing 42, 1778-1790.

Mnih, V., Hinton, G.E., 2010. Learning to Detect Roads in High-Resolution Aerial Images, Computer Vision - ECCV 2010 - European Conference on Computer Vision, Heraklion, Crete, Greece, September 5-11, 2010, Proceedings, pp. 210-223.

Nogueira, K., Penatti, O.A.B., Santos, J.A.D., 2017. Towards better exploiting convolutional neural networks for remote sensing scene classification. Pattern Recognition 61, 539-556.

Ojala, T., Pietikainen, M., Maenpaa, T., 2002. Multiresolution gray-scale and rotation invariant texture classification with local binary patterns. IEEE transactions on pattern analysis and machine intelligence 24, 971-987.

Oliva, A., Torralba, A., 2001. Modeling the Shape of the Scene: A Holistic Representation of the Spatial Envelope. International Journal of Computer Vision 42, 145-175.

Penatti, O.A.B., Nogueira, K., Santos, J.A.D., 2015. Do deep features generalize from everyday objects to remote sensing and aerial scenes domains?, IEEE Conference on Computer Vision and Pattern Recognition Workshops, pp. 44-51.

Redmon, J., Divvala, S., Girshick, R., Farhadi, A., 2016. You Only Look Once: Unified, Real-Time Object Detection. 779-788.

Ren, S., He, K., Girshick, R., Sun, J., 2015. Faster R-CNN: Towards Real-Time Object Detection with Region Proposal Networks, International Conference on Neural Information Processing Systems, pp. 91-99.

Rezić, A., 2011. Ant colony optimization. Alphascript Publishing volume 28, 1155-1173.

Rice M T, P.F.I., Mulhollen A P, et al., 2012. Crowdsourced Geospatial Data: A report on the emerging phenomena of crowdsourced and user-generated geospatial data. GEORGE MASON UNIV FAIRFAX VA.

Rottensteiner, F., Sohn, G., Jung, J., Gerke, M., Baillard, C., Benitez, S., Breitkopf, U., 2012. The Isprs Benchmark on Urban Object Classification and 3d Building Reconstruction. I-3, 293-298.

Salberg, A.-B., 2015. Detection of seals in remote sensing images using features extracted from deep convolutional neural networks, Geoscience and Remote Sensing Symposium.
Simonyan, K., Zisserman, A., 2014a. Two-Stream Convolutional Networks for Action Recognition in Videos. Advances in Neural Information Processing Systems 1, 568-576.
Simonyan, K., Zisserman, A., 2014b. Very Deep Convolutional Networks for Large-Scale Image Recognition. Computer Science.
Sun, Y., Wang, X., Tang, X., 2013. Hybrid Deep Learning for Face Verification, international conference on computer vision, pp. 1489-1496.
Sun, Y., Wang, X., Tang, X., 2014a. Deep Learning Face Representation by Joint Identification-Verification. 27, 1988-1996.
Sun, Y., Wang, X., Tang, X., 2014b. Deep Learning Face Representation from Predicting 10,000 Classes. 1891-1898.
Sun, Y., Wang, X., Tang, X., 2015. Deeply learned face representations are sparse, selective, and robust, Computer Vision and Pattern Recognition, pp. 2892-2900.
Swain, M.J., Ballard, D.H., 1991. Color indexing. International Journal of Computer Vision.
Szegedy, C., Liu, W., Jia, Y., Sermanet, P., Reed, S., Anguelov, D., Erhan, D., Vanhoucke, V., Rabinovich, A., 2014. Going deeper with convolutions. 1-9.
Taigman, Y., Yang, M., Ranzato, M.A., Wolf, L., 2014. DeepFace: Closing the Gap to Human-Level Performance in Face Verification, IEEE Conference on Computer Vision and Pattern Recognition, pp. 1701-1708.
Vogelmann, J.E., 2001. Completion of the 1990s National Land Cover Data Set for the Conterminous United States from Landsat Thematic Mapper Data and Ancillary Data Sources. Photogrammetric Engineering & Remote Sensing 67, 650-655.
Wu, H., Liu, B., Su, W., Zhang, W., Sun, J., 2016. Hierarchical Coding Vectors for Scene Level Land-Use Classification. Remote Sensing 8, 436.
Xia, G.S., Hu, J., Hu, F., Shi, B., Bai, X., Zhong, Y., Zhang, L., 2016. AID: A Benchmark Dataset for Performance Evaluation of Aerial Scene Classification.
Xian, G., Homer, C., Dewitz, J., Fry, J., Hossain, N., Wickham, J., 2011. Change of impervious surface area between 2001 and 2006 in the conterminous United States. Photogrammetric Engineering & Remote Sensing 77, 758-762.
Yan, X., Chang, H., Shan, S., Chen, X., 2014. Modeling Video Dynamics with Deep Dynencoder. Springer International Publishing.
Yang, Y., Newsam, S., 2010. Bag-of-visual-words and spatial extensions for land-use classification, ACM Sigspatial International Symposium on Advances in Geographic Information Systems, Acm-Gis 2010, November 3-5, 2010, San Jose, Ca, Usa, Proceedings, pp. 270-279.
Yao, X., Han, J., Cheng, G., Qian, X., Guo, L., 2016. Semantic Annotation of High-Resolution Satellite Images via Weakly Supervised Learning. IEEE Transactions on Geoscience and Remote Sensing 54, 3660-3671.
Yao, X., Han, J., Zhang, D., Nie, F., 2017. Revisiting Co-Saliency Detection: A Novel Approach based on Two-stage Multi-view Spectral Rotation Co-clustering. IEEE Transactions on Image Processing 26, 3196-3209.
Zeng, X., Ouyang, W., Yang, B., Yan, J., Wang, X., 2016. Gated Bi-directional CNN for Object Detection.

Zhao, B., Zhong, Y., Xia, G., Zhang, L., 2016a. Dirichlet-Derived Multiple Topic Scene Classification Model for High Spatial Resolution Remote Sensing Imagery. IEEE Transactions on Geoscience and Remote Sensing 54, 2108-2123.

Zhao, B., Zhong, Y., Zhang, L., Huang, B., 2016b. The Fisher Kernel Coding Framework for High Spatial Resolution Scene Classification. Remote Sensing 8, 157.

Zhou, B., Khosla, A., Lapedriza, A., Oliva, A., Torralba, A., 2016. Learning Deep Features for Discriminative Localization, Computer Vision and Pattern Recognition, pp. 2921-2929.

Zhu, Q., Zhong, Y., Zhao, B., Xia, G., Zhang, L., 2016. Bag-of-Visual-Words Scene Classifier With Local and Global Features for High Spatial Resolution Remote Sensing Imagery. IEEE Geoscience and Remote Sensing Letters 13, 747-751.